\documentclass[11pt]{article}
\usepackage{graphicx}
\usepackage{fullpage}
\usepackage{amsthm}
\usepackage{amsmath}
\usepackage{amssymb}
\usepackage{amsfonts}
\usepackage{xcolor}
\usepackage{booktabs}
\usepackage[round]{natbib}
\usepackage{hyperref}
\usepackage{cleveref}
\usepackage{tikz}
\title{
{Rural Readiness: Domain Generalization with Adversarially Chosen Domains}
}
\author{Cynthia Dwork \\ \small Harvard University \and Andisheh Ghasemi \\ \small Northeastern University \and Lunjia Hu \\ \small Northeastern University}
\date{}

\newtheorem{theorem}{Theorem}
\newtheorem{lemma}[theorem]{Lemma}
\newtheorem{corollary}[theorem]{Corollary}
\newtheorem{definition}{Definition}
\newtheorem{remark}{Remark}
\newtheorem{proposition}[theorem]{Proposition}
\crefname{corollary}{corollary}{corollaries}
\Crefname{corollary}{Corollary}{Corollaries}
\newcommand{\E}{\mathbb E}
\newcommand{\cL}{\mathcal L}
\newcommand{\cF}{\mathcal F}
\newcommand{\cD}{\mathcal D}
\newcommand{\sps}[1]{^{(#1)}}
\newcommand{\err}{\mathrm{err}}
\newcommand{\kl}{\mathrm{kl}}
\newcommand{\cG}{\mathcal{G}}
\newcommand{\cC}{\mathcal{C}}
\newcommand{\Gdim}{\mathrm{Gdim}}
\newcommand{\cX}{\mathcal{X}}

\DeclareMathOperator*{\argmax}{arg\,max}
\DeclareMathOperator*{\argmin}{arg\,min}

\begin{document}

\maketitle

\begin{abstract}
Predictive models trained at well-resourced data-rich institutions often fail when deployed in new populations. Motivated by the goal of building clinical diagnosis and decision-making algorithms for data-scarce rural populations using the general knowledge learned from data-rich hospitals, we formulate the fundamental question of \emph{domain generalization with adversarially chosen domains}, which asks a learner to perform on any unseen target domain nearly as well as the worst hypothesis consistent with the evidence from the observed source domains.

Assuming only that the hypothesis class contains a ``universal'' predictor with small error on all $k$ source domains, we characterize the sample complexity of this objective up to logarithmic factors. Letting $d$ denote the VC dimension of the hypothesis class, we find, somewhat surprisingly, that the sample complexity reduces significantly from $\Omega(dk)$ to $O(d)$ under a small relaxation of the learning goal in which, roughly speaking, we allow the learned model to compete with hypotheses that perform well on 99\% instead of 100\% of the source domains. By applying this insight to the meta-distributional setting, in which source domains are selected according to a distribution on distributions, we obtain a corresponding improvement over known sample complexity upper bounds. We also study the effect of data access to the target domain and show that a small labeled sample from the target domain does not improve the worst-case sample complexity on the source domains.

\end{abstract}

\section{Introduction}
Consider the task of creating a diagnostic tool that, given information about a patient such as lab results and vitals, outputs a score in $[0,1]$, with higher numbers viewed as a positive diagnosis.  Suppose well-funded, data-rich medical research centers have developed diagnostic tools that perform well on the local populations on which they were trained.  It is well known among medical researchers that a scoring function developed in one facility can be significantly less effective at other facilities~\citep{rvr, patil2018training}.
This can arise from technical factors such as the methods and batches of chemicals used for specimen collection, as well as population differences involving diets, environmental exposures, and genetic factors. These factors can change the joint probability distribution over patient information and ground-truth diagnoses; we call such a distribution a \emph{domain}.
The \emph{source domains} on which a diagnostic tool is trained may differ from the \emph{target domains} on which it is evaluated.
The resulting performance degradation from source to target is known as the {\em Harvard--Stanford problem}: letting $D_H$ and $D_S$ denote the distributions on which Harvard and Stanford predictors $f_H$ and $f_S$ are trained, respectively, the cross-domain errors $\err_{D_S}(f_H)$ and $\err_{D_H}(f_S)$ can be higher than the within-domain errors $\err_{D_H}(f_H)$ and $\err_{D_S}(f_S)$.
As long as in practice each predictor is used only for the distribution on which it is trained, there is no immediate problem.

Unfortunately, not all populations live near well-resourced, data-rich, cutting-edge medical research centers. At the same time, it is believed that human populations are not all {\em so} different from one another.
This led~\cite{rvr} to pose the following question: assuming the existence of an ``$\alpha$-universal'' predictor, meaning, informally, that it has error at most $\alpha$ on all populations of interest, for some small $\alpha>0$ (although it might not be optimal for any single population), can we leverage samples from the observed populations to {\em construct} a $\beta$-universal predictor, for some $\beta\ge \alpha$? We call this goal {\em achieving rural readiness}; it is closely related to domain generalization. Our formulation without assumptions relating different populations is one of our principal contributions.   As noted in~\cite{rvr}, rural readiness ``provides a new dimension of fairness, transferring the benefit of federal research dollars from preeminent bench to geographically remote bedside.''

A simple example shows that, absent assumptions relating populations, achieving absolute rural readiness using data collected from even a large set of observed populations can be impossible: if all observed populations are identical, and if these identical distributions differ markedly from an unobserved population, then there is little reason to think lessons learned from the observed populations will generalize to the new population. This suggests a competitive notion capturing the goal of building a scoring function that generalizes to unseen target domains ``as well as possible'' {\em given the source domains that have been observed}. We call this {\em conditional} rural readiness, or simply {\em rural readiness}, since the absolute version is impossible. We describe the competitive learning goal as follows.

Fix a hypothesis class $\cF \subseteq\{f:\cX\rightarrow \{0,1\}\}$, where $\cX$ is an instance space (in our discussion, its elements describe patients).
Although in the motivating discussion, we assume there is an $\alpha$-universal predictor that has error at most $\alpha$ on all populations of interest, in our formal treatment we assume only that $\mathcal F$ contains a predictor $f^\star$ with error at most $\alpha$ on each of the observed source domains.
We call any hypothesis $f^\star\in \cF$ satisfying this property $\alpha$-universal.
This weaker assumption suffices for
our competitive learning goal.
In particular, $f^\star$ need not have low error on the unobserved target domain.\footnote{As will become clear, this is analogous to \emph{agnostic} learning, as opposed to \emph{realizable} learning.}
Throughout the paper, our upper bounds hold under this source-only assumption, whereas our lower bounds hold even in the easier setting where $f^\star$ has zero error on the target domain.

Each observed source domain eliminates hypotheses whose error exceeds $\beta$ on that domain. The survivors form a {\em benchmark set}. Without assumptions relating sources to targets, the source data provide no basis for preferring one surviving hypothesis over another on an unseen target $D$. We therefore require the learned classifier to perform, up to an additive slack, at least as well as the worst-performing benchmark hypothesis on $D$. If the sources are informative, every benchmark hypothesis performs well on the target domain; if they are not, the benchmark permits large target error and the guarantee appropriately becomes vacuous.

We also consider a larger, relaxed benchmark. If a function $f\in\cF$ performs well on, say, $99\%$ rather than $100\%$ of the observed sources, this may still be evidence that it may perform well on target distributions. The relaxed benchmark therefore permits a hypothesis to fail on a $\gamma$ fraction of the sources. It contains the strict benchmark, which permits no failures, but may include additional members of~$\cF$.

We study the sample complexity of achieving (conditional) rural readiness, assuming only the existence of an $\alpha$-universal predictor in $\cF$, obtaining upper and lower bounds that match up to logarithmic factors for both the strict and relaxed benchmark sets.  A key insight from our analysis is that, perhaps surprisingly, the sample complexity for the relaxed version is dramatically smaller than for the strict version.

Rural readiness is a domain generalization problem in which the source and target domains can all be chosen adversarially.  The formulation has applications beyond the medical scenario that originally motivated our research, so in our technical results we use the term {\em adversarial domain generalization} for this learning goal.\footnote{In contrast, the literature also uses {\em adversarial} to describe a family of representation-learning techniques. Domain-adversarial neural networks (DANNs) were introduced for domain adaptation~\citep{ganin2016domain}, whereas {\em adversarial domain generalization} (ADG) refers to related methods that adversarially align features across source domains~\citep{zhu2022localized}.}

\subsection{Summary of Contributions}

We formalize the problem of achieving {\em (conditional) rural readiness} without assumptions relating the source and target distributions, and characterize its sample complexity.
Our key modeling contribution is a competitive formulation with an appropriate choice of benchmark that formalizes the goal of doing as well as possible given what has been observed.

\paragraph{Adversarial Domain Generalization.}
We formalize the objective of conditional rural readiness using the \emph{adversarial domain generalization} task in \Cref{sec:setup}. In this task, the goal is to design a learning algorithm that collects data from $k$ adversarially chosen source domains $\cD = (D_1,\ldots,D_k)$, and outputs a model that achieves competitive error on unseen target domains. We make the $\alpha$-universality assumption: there exists an unknown $f^\star$ from a hypothesis class $\cF$ achieving error at most $\alpha$ on every source domain. The competitive error goal is defined with respect to the benchmark set $\cF_\cD(\beta)$ consisting of hypotheses $f\in \cF$ whose error on every source domain is at most $\beta$, for $\beta \ge \alpha$. We say a model achieves competitive error on a target domain if its error is at most the maximum error of a benchmark $f\in \cF_\cD(\beta)$ on that target domain, plus $\varepsilon$. We also consider the larger, relaxed benchmark set $\cF_\cD\sps \gamma(\beta)$ consisting of hypotheses $f\in \cF$ that achieve error at most $\beta$ on at least a $(1-\gamma)$ fraction of the $k$ source domains.

\paragraph{Near-Matching Bounds for the Strict Benchmark.}
We prove near-matching upper and lower bounds on the sample complexity of adversarial domain generalization in \Cref{sec:lb,sec:ub}. We start with the setting with the strict benchmark set $\cF_\cD(\beta)$:

\begin{theorem}[informal, see \Cref{cor:lb-vc} and \Cref{thm:ub}]
\label[theorem]{thm:informal-strict}
    Assume $0\le \alpha<\beta<1/4$, $0\le\varepsilon<1/4$, $0<\delta<1/4$, and $d\ge10\log k$. In the worst case over hypothesis classes of VC dimension at most $d$, the sample complexity of adversarial domain generalization is, up to logarithmic factors,
    \[
    \frac{\beta}{(\beta - \alpha)^2}\cdot k(d + \log(1/\delta)).
    \]
\end{theorem}

The $\Omega(kd)$ lower bound is in sharp contrast to recent results on multi-distribution learning~\cite{DBLP:journals/jacm/ZhangZCDL25}, where the sample complexity for finding a predictor that performs well on $k$ given distributions is only an additive $k+d$.  When $\beta>2\alpha$, these better bounds apply to our problem if the family $\cF$ satisfies a strong {\em majority-closure} condition: for every finite multiset $f_1,\dots,f_T \in \cF$, the function that assigns to $x$ the majority value assigned to $x$ by $f_1,\dots,f_T$ is also in $\cF$; see Appendix~\ref{app:mdl-closure}.

A strength of our sample complexity lower bound in \Cref{thm:informal-strict} is that it holds for an easier task. Our adversarial domain generalization task requires the learner to output a single model that achieves a competitive error guarantee for all target domains. Our lower bound holds even if the error guarantee is only required for a single target domain that depends solely on the source domains and is independent of the sampled data points and the learner's randomness. Moreover, that single target domain is fully realizable: the $\alpha$-universal hypothesis $f^\star\in \cF$ is the only member of the benchmark family $\cF_\cD(\beta)$ and it achieves zero error on the target domain.

To prove the lower bound, we carefully construct a family of hard instances, where solving the adversarial domain generalization problem requires the learner to fully identify $f^\star$.
This construction allows us to use an adversarial domain generalization learner to solve \emph{hidden-coordinate identification problems}. These problems isolate the key information-theoretic insight in our proof. We describe these problems in \Cref{sec:hidden} and apply standard information-theoretic tools (e.g.\ Fano's inequality and the information processing inequality for KL divergences) to prove lower bounds for them in \Cref{app:hidden-coordinate-proof}. This allows us to present the rest of the proof in a modular way in \Cref{sec:lb}.

To prove the sample complexity upper bound in \Cref{thm:informal-strict}, we design an algorithm that achieves a stronger goal: with probability at least $1-\delta$, the algorithm outputs a member of the benchmark set $\cF_\cD(\beta)$. Since our competitive error goal is with respect to the worst hypotheses in the benchmark set, any member of the benchmark set is a valid solution to the adversarial domain generalization problem. Our algorithm uses source data points to estimate the error of hypotheses $f\in \cF$ on every source domain, in order to rule out all hypotheses not in the benchmark set $\cF_\cD(\beta)$. The $\alpha$-universality assumption ensures that the algorithm does not rule out all hypotheses, allowing the algorithm to choose its output from the remaining hypotheses. Obtaining a tight upper bound requires carefully designing the criterion for ruling out hypotheses, which we describe in more detail in \Cref{sec:ub}.

\paragraph{Near-Matching Bounds for a Relaxed Benchmark.}
Our results make no distributional assumptions, but they do reflect a distributional {\em optimism}: the observed distributions are sufficiently informative to teach us how to make predictions for as-yet-unobserved distributions. %
Our relaxed benchmark captures even greater optimism, specifically, that the information needed to make good predictions on an unobserved distribution can be learned even while disregarding a small fraction of the observed sources.  This could happen, for example, if we believe some observed sources may be pathological or anomalous. %

If there is reason to be strongly optimistic in the above sense, we can aim for a relaxed benchmark
 class $\cF_\cD\sps \gamma(\beta)$ in which we retain candidates from $\cF$ that may fail to be accurate on up to a small constant fraction $\gamma$ of the observed domains.
 This fully eliminates the linear dependence on $k$ in \Cref{thm:informal-strict}:
 \begin{theorem}[informal, see \Cref{cor:lb-vc,thm:ub}]
 \label[theorem]{thm:informal-relaxed}
     In the setting of \Cref{thm:informal-strict}, assume $\gamma\in[1/k,1/4]$. In the worst case over hypothesis classes of VC dimension at most $d$, the source-sample complexity of $\gamma$-relaxed adversarial domain generalization is, up to logarithmic factors,
     \[
     \frac{\beta}{(\beta - \alpha)^2}\cdot \frac 1{\gamma} \cdot(d + \log(1/\delta)).
     \]
 \end{theorem}
Interestingly, the sample complexity bound in \Cref{thm:informal-relaxed} is independent of $k$, demonstrating the power of the benchmark relaxation. As an application of this result, we use it to achieve improved sample complexity for the domain generalization problem studied by \citet{dhs}.

We prove the lower bound in \Cref{thm:informal-relaxed} by reducing to the strict-benchmark bound in \Cref{thm:informal-strict}. Set $r=\lfloor\gamma k\rfloor+1$ and $B=\lfloor k/r\rfloor$, start from a strict instance with $B$ sources, and repeat each source at least $r$ times. A hypothesis that violates the threshold on one original source then violates it on more than $\gamma k$ copies. The relaxed benchmark of the repeated instance is therefore exactly the strict benchmark of the original instance.

Since the sample complexity upper bound in \Cref{thm:informal-relaxed} is independent of $k$, our algorithm need not collect data from every source domain when $k$ is large. The key idea is to sample a subset of the source domains and collect data only from those sampled domains. The size of this subset needs to be chosen carefully to obtain a tight bound. We describe the details in \Cref{sec:ub}.

\noindent
\paragraph{Improved Upper Bounds for Meta-Generalization.} There is a large literature on domain generalization when the observed distributions are themselves drawn from a distribution over distributions. Dwork, Hu, and Shao~\citep{dhs} model this in the PAC framework, introduce the {\em domain shattering dimension}, and give tight bounds on the number of randomly sampled domains needed to achieve absolute rural readiness.
In this setting, we prove that every member of the relaxed benchmark generalizes to new domains from the meta-distribution. For constant parameters, combining this transfer result with our learner gives total labeled sample complexity $\widetilde O(\mathrm{VC}(\cF))$, independent of $k$, rather than the $\widetilde O(k\,\mathrm{VC}(\cF))$ used by the min--max ERM implementation analyzed in~\cite{dhs}. This saving concerns labeled examples: the number $k$ of observed domains must still scale with the domain shattering dimension, and the transferred bad-domain probability includes an additive $\gamma$.

\paragraph{Data Access to the Target Domain.} Our results above assume that no target data are available. Perhaps surprisingly, a small labeled target sample does not improve the worst-case source-sample complexity: the source lower bound persists until the target sample is large enough for target-only learning.

\begin{theorem}[informal, see \Cref{thm:m-target-lower}]
\label[theorem]{thm:informal-target}
    Given data access to the target domain, the source-domain lower bounds in \Cref{thm:informal-strict,thm:informal-relaxed} continue to hold up to constant factors whenever the number of labeled target examples is smaller than a constant multiple of
    \begin{equation}
    \label{eq:informal-target}
    \frac{d + \log(1/\delta)}{\varepsilon^2}
    \end{equation}
    data points from the target domain, where $\varepsilon$ is the additive slack in the competitive error goal.
\end{theorem}
Conversely, standard agnostic learning uses $O((d+\log(1/\delta))/\varepsilon^2)$ labeled target examples and no source examples.
We also prove a realizable counterpart: even when the benchmark has zero
target error, the source lower bound persists below the
$(d+\log(1/\delta))/\varepsilon$ target-sample scale; see
\Cref{thm:target-access-epsilon}. This matches the
usual realizable target-only sample complexity. These results hold in the relaxed-benchmark setting by the same repeated-source reduction (see \Cref{cor:target-access-vc}).
\subsection{Related Work}
\label[subsection]{sec:related_work}
Speaking intuitively, both similarities and differences across observed populations can be leveraged to achieve rural readiness. Similarities help isolate invariant mechanisms that reliably transfer to unseen populations, whereas differences alert the learner to spurious or domain-specific features that must be avoided.

Approaches exploiting similarities typically seek stable causal mechanisms via invariant risk minimization~\citep{arjovsky2019invariant}. Conversely, other approaches explicitly exploit domain or group differences, for example through adversarial domain discrimination~\citep{rvr}, distributionally robust optimization across environments~\citep{sagawa2019distributionally}, or adversarial fairness and transfer objectives~\citep{madras2018learning}. Theoretical work by \citet{rosenfeld2020risks} and \citet{gouk2022limitations} identifies specific formal settings in which invariant or general-purpose domain-generalization methods cannot improve fundamentally over empirical risk minimization (pooling-ERM). Empirical work, including \citet{teterwak2025erm++}, likewise emphasizes that carefully tuned ERM-style baselines can be highly competitive against specialized domain generalization methods on standard benchmarks. For comprehensive overviews of domain generalization and transfer learning, see~\citet{zhou2022domain,wang2022generalizing}.

\paragraph{Multi-Distribution and Collaborative Learning (Observed Sources).}
A rich line of work investigates learning predictors evaluated exclusively across $k$ known source distributions. Initiated by \citet{blum2017collaborative}, collaborative PAC learning characterizes the sample complexity of finding a shared classifier that simultaneously satisfies all participating domains, with matching logarithmic-overhead bounds established in relevant collaborative-PAC regimes by \citet{nguyen2018improved} and \citet{chen2018tight}. Recent work has studied demand-driven on-demand query allocation~\citep{haghtalab2022demand}, formalized open questions regarding sample complexity for VC classes~\citep{awasthi2023sample}, settled the minimax sample complexity of multi-distribution learning~\citep{peng2024sample,DBLP:journals/jacm/ZhangZCDL25}, and developed derandomized collaborative algorithms~\citep{larsen2024derandomizing}. Extending multi-distribution learning to noisy settings, \citet{DBLP:conf/colt/HanashiroSJ26} show that in personalized multi-distribution learning, bounded label noise can preclude fast $1/\epsilon$-type rates, leading instead to $k/\epsilon^2$-type dependence or multiplicative penalties under stronger benchmarks. These frameworks assess performance strictly \emph{in-sample} on observed distributions $D_1, \ldots, D_k$; in contrast, our goal is out-of-domain generalization to an unobserved, potentially adversarial target distribution $D'$.

\paragraph{Assumptions on Source--Target Similarity and Divergences.}
To generalize to an unobserved target domain, the dominant paradigm in domain adaptation and domain generalization is to restrict the discrepancy between source and target distributions through explicit geometric, statistical, or algebraic bounds. In domain adaptation, classical theory bounds target risk using discrepancy measures such as $\mathcal{H}\Delta\mathcal{H}$-divergence~\citep{ben2010theory}, while target-specific inference often uses propensity-score reweighting~\citep{kim2022universal}. In domain generalization without target data, early theoretical frameworks control transfer to an unseen target by bounding distribution discrepancies in an RKHS, either by embedding domain marginal distributions~\citep{blanchard2011generalizing} or minimizing maximum mean discrepancy to learn domain-invariant representations~\citep{muandet2013domain}. Alternatively, agnostic multi-distribution learning restricts the target domain to lie within the convex mixture simplex of the observed source distributions~\citep{mohri2019agnostic}, while group-theoretic approaches bound worst-case generalization by assuming target distributions are generated by known transformation-group symmetries acting on the source data~\citep{shao2022theory,montasser2024transformation}. In contrast to all these approaches, our adversarial domain generalization framework operates under zero target data and imposes no divergence bounds, distribution-distance constraints, mixture simplex assumptions, or transformation symmetries between the observed sources and the unseen target.

\paragraph{Causal Mechanisms, Drift, and Representation Learning.}
Beyond direct distance bounds between domains, related lines of work generalize across environments by isolating invariant causal mechanisms or shared feature representations. Causal frameworks rely on structural causal models and causal DAGs to identify invariant conditional distributions~\citep{scholkopf2012causal,zhang2015multi,gong2016domain}, while other works address transfer under posterior and decision-rule drift~\citep{maity2024linear,zhu2024label,zhu2025domain,wang2025transfer}, where $P(Y|X)$ shifts across domains. In empirical representation learning, adversarial domain generalization methods align feature representations across source domains~\citep{zhu2022localized}, adapting gradient-reversal techniques originally developed for source-target alignment in domain adaptation~\citep{ganin2016domain}. A parallel theoretical line in multi-task and representation learning guarantees transfer by assuming that all tasks share a common low-dimensional linear representation or subspace~\citep{balcan2015efficient,maurer2016benefit,kong2020meta,tripuraneni2020theory,tripuraneni2021provable,du2020few,thekumparampil2021sample,collins2021exploiting}. Recent advances in this setting establish optimal subspace recovery under heterogeneous noise~\citep{duchi2022subspace} and in extreme regimes with only constant samples per task~\citep{aliakbarpour2024metalearning}. Rather than restricting the data-generating process to specific parametric forms, causal graphs, or low-dimensional linear representations, we operate solely under the $\alpha$-universality assumption over general hypothesis classes $\cF$.

\paragraph{Meta-Learning and Meta-Distributional Generalization.}
Meta-learning typically models domains as independent draws from an underlying meta-distribution $\mathcal{M}$. Several formulations evaluate average performance over domains drawn from a meta-distribution~\citep{garg2021learn}; in some cases this reduces to learning under the induced mixture, while other formulations retain explicit domain-level structure. Recent work also investigates fundamental ERM principles and sample requirements in meta-learning~\citep{alon2024erm}, or studies online meta-learning for fast parameter initialization~\citep{khodak2019adaptive,khodak2023meta,balcan2021learning}. When bounding uniform error across unseen domains supported on $\mathcal{M}$, prior work relies on domain complexity measures such as the domain-shattering dimension introduced by \citet{dhs}. While the min-max ERM algorithm of \citet{dhs} achieves optimal domain complexity ($k$), its literal implementation requires estimating observed-domain empirical errors uniformly to high precision, incurring a total sample cost of $\tilde{O}(k \cdot \mathrm{VC}(\cF))$. By framing meta-generalization through our relaxed benchmark, we reduce this total hospital-side sample burden to $\tilde{O}(\mathrm{VC}(\cF))$ for constant parameters, independent of $k$, while maintaining strong out-of-domain guarantees.
\subsection{Paper Organization}
The rest of the paper is organized as follows.
\Cref{sec:setup} formalizes adversarial domain generalization, universality, and the strict and relaxed benchmarks. \Cref{sec:lb} proves the strict and relaxed sample-complexity lower bounds, while \Cref{sec:ub} gives near-matching upper bounds via subset certificates and random-domain batches. \Cref{sec:target-samples} studies how labeled target samples change the lower bounds, and \Cref{sec:mdl-generalization} applies the relaxed benchmark to meta-distribution generalization. \Cref{sec:tools} collects the information-theoretic and VC tools, \Cref{app:subset-certificate} proves the two upper bounds, \Cref{app:hidden-coordinate-proof} proves the hidden-coordinate lemmas, \Cref{app:epsilon-target-access} gives the detailed target-access lower bounds, and Appendix~\ref{app:mdl-closure} develops improved guarantees under majority closure.

\section{Problem Setup: Adversarial Domain Generalization}
\label[section]{sec:setup}
In this section, we formally define the \emph{adversarial domain generalization} problem, starting from two key concepts: the universality assumption and the benchmark set.
\subsection{The Universality Assumption}
Let $\mathcal{X}$ be an instance space and let $\mathcal{Y} = \{0,1\}$ be the binary label space. A
hypothesis class is a family $\cF$ of binary classifiers $f:\cX\to \{0,1\}$.
For a distribution $D$ on $\mathcal{X} \times \mathcal{Y}$ and a classifier
$f : \mathcal{X} \to \{0,1\}$, we define
\[
    \err_D(f) := \Pr_{(x,y)\sim D}[f(x) \neq y].
\]
Suppose we have data access to $k$ source distributions
$\cD := (D_1, \ldots, D_k)$.  (Note that $\cD$ is a list, not a distribution.)
For example, each $D_i$ represents the distribution of patients $x$ in a specific hospital with their corresponding labels $y$.
The learner may request labeled examples from any source; each request to
$D_i$ returns a fresh i.i.d.\ draw from that distribution.
Following the terminology in domain generalization, we call these distributions \emph{source domains}.

We make the following $\alpha$-universality assumption on the hypothesis class $\cF$ and the source distributions $D_1,\ldots,D_k$:
\begin{definition}[$\alpha$-universality]
\label[definition]{def:alpha-universality}
    We say a hypothesis class $\cF$ satisfies $\alpha$-universality on distributions $D_1, \ldots, D_k$ if there exists $f^\star\in \cF$ such that
    \[
    \err_{D_i}(f^\star) \le \alpha \quad \text{for all $i = 1,\ldots,k$.}
    \]
\end{definition}
We think of $\alpha$ as a small positive constant in $[0,1]$, say, $\alpha = 0.1$ for concreteness.
The $\alpha$-universality condition requires that some $f^\star\in \cF$ performs reasonably well on all the source domains. Ideally, $f^\star$ would capture some universal scientific knowledge that holds across all realistic patient distributions, but our results make no such assumption. That said, the classifier $f^\star$ need not be optimal on any specific source domain $D_i$: another $D_i$-specific classifier $f_{D_i}\in \cF$ may achieve significantly smaller error on $D_i$.

\subsection{The Benchmark Set}

Our goal is to use data collected from the source domains to learn a universal classifier $\hat f:\cX\to \{0,1\}$ that performs well on new \emph{target} domains $D'$. We study a challenging setting:
\begin{enumerate}
    \item We do not make any explicit assumption on the similarity between the source domains and the target domain. This is in contrast to prior work on domain generalization that relies on distance or divergence bounds between the source and the target (see \emph{Assumptions on Source--Target Similarity and Divergences} in Section~\ref{sec:related_work}). This is also different from meta-distributional models in which domains are sampled i.i.d.~\citep{dhs} (see \emph{Meta-Learning and Meta-Distributional Generalization} in Section~\ref{sec:related_work}).
    \item We train the classifier $\hat f$ without using any data from the target distribution $D'$.
\end{enumerate}

\begin{figure}[ht]
\centering
\begin{tikzpicture}[scale=0.95, every node/.style={transform shape}]
  \draw[thick, rounded corners=12pt, fill=gray!4, draw=gray!40] (-5.2,-4.2) rectangle (5.2,4.2);
  \node[anchor=north west, font=\small\bfseries, color=gray!70!black] at (-4.9,3.9) {Hypothesis Space $\cF$};

  \foreach \angle/\name in {90/1, 162/2, 234/3, 306/k-1, 18/k} {
    \draw[rotate around={\angle:(0,0)}, fill=blue!10, fill opacity=0.25, draw=blue!70!black, thick]
      (0,0.85) ellipse (1.35cm and 2.45cm);
    \node[font=\footnotesize\bfseries, color=blue!80!black] at (\angle:3.3) {$D_{\name}$};
  }

  \draw[fill=orange!35, fill opacity=0.6, draw=orange!90!black, thick, densely dashed] (0,0) circle (1.15cm);

  \draw[fill=green!40, fill opacity=0.75, draw=green!60!black, very thick] (0,0) circle (0.55cm);

  \node[font=\footnotesize\bfseries, color=orange!90!black] at (0, 0.82) {$\cF_\cD(\beta)$};
  \node[font=\tiny\bfseries, color=green!40!black] at (0, 0.22) {$\cF_\cD(\alpha)$};

  \filldraw[black] (0,-0.18) circle (2.2pt) node[below, font=\small\bfseries] {$f^\star$};
  \filldraw[red!80!black] (0.78,-0.18) circle (2.2pt) node[right, font=\small\bfseries] {$f'$};

\end{tikzpicture}
\caption{Geometry of the benchmark $\cF_\cD(\beta)$.
Each petal represents the hypotheses whose population error is at most
$\beta$ on one source $D_i$.  Their intersection is
$\cF_\cD(\beta)$; the inner region
$\cF_\cD(\alpha)$ contains the universal predictor $f^\star$
from \Cref{def:alpha-universality}.  The classifier
$f'\in\argmax_{f\in\cF_\cD(\beta)}
\err_{D'}(f)$ is the benchmark member with largest
target error.  Both $f^\star$ and $f'$ satisfy the relevant source constraints,
but without assumptions relating source and target domains, source evidence
alone gives no smaller uniform target guarantee than the error of this worst
benchmark member.
}
\label{fig:sunflower-benchmark}
\end{figure}

In this challenging setting without any data from the unconstrained target distribution $D'$, it is impossible to always achieve a non-trivial error guarantee on it. We aim for the relaxed goal of achieving competitive error with respect to a benchmark set of classifiers:

\begin{definition}[Benchmark set]
\label[definition]{def:benchmark}
    Given a hypothesis class $\cF$, source distributions $\cD = (D_1,\ldots,D_k)$, and an error parameter $\beta\in [0,1]$, we define the \emph{benchmark set}
    \[
        \cF_\cD(\beta) := \Bigl\{f\in\cF : \max_{i\in[k]} \err_{D_i}(f)
        \le \beta\Bigr\}.
    \]
    When $\beta$ is understood from context we may simply write $\cF_\cD$.
\end{definition}

The benchmark set $\cF_\cD(\beta)$ consists of classifiers $f\in \cF$ that perform reasonably well on all source distributions. Assuming $\alpha$-universality for $\alpha\le \beta$, it is clear by definition that the $\alpha$-universal classifier $f^\star$ belongs to the benchmark set $\cF_\cD(\beta)$.

To see the intuition behind \Cref{def:benchmark}, let us consider the special case where $\beta = \alpha$. In this case, $\cF_\cD(\alpha)$ is exactly the set of candidates for $f^\star$ if we fully observed the source distributions $D_1,\ldots,D_k$. That is, even if we have full knowledge of $D_1,\ldots,D_k$, the only information we can learn about $f^\star$ is that it is a member of $\cF_\cD(\alpha)$.
If we take an arbitrary $\hat f \in \cF_\cD(\alpha)$ as a guess for $f^\star$, we immediately obtain the following distribution-free guarantee on a new target distribution $D'$:
\begin{equation}
\label{eq:benchmark-error}
\err_{D'}(\hat f) \le \max_{f\in \cF_\cD(\alpha)} \err_{D'}(f).
\end{equation}
This motivates \eqref{eq:benchmark-error} as a conservative competitive guarantee based only on the source distributions, even if those distributions were known exactly.

We also consider the following $\gamma$-relaxed benchmark set:
\begin{definition}[Relaxed benchmark set]
\label[definition]{def:relaxed-benchmark}
    In the setting of \Cref{def:benchmark}, for $\gamma \in [0,1]$, define
    \[
        \cF_\cD^{(\gamma)}(\beta)
        := \Bigl\{f \in \cF :
        \bigl|\{i \in [k] : \err_{D_i}(f) > \beta\}\bigr|
        \le \gamma k\Bigr\}.
    \]
    Equivalently, $f \in \cF_\cD^{(\gamma)}(\beta)$ may violate the threshold $\beta$ on
    at most $\lfloor\gamma k\rfloor$ sources.
\end{definition}
It is clear that $\cF_\cD(\beta) = \cF_\cD\sps 0(\beta)$ and $\cF_\cD\sps {\gamma_1}(\beta) \subseteq \cF_\cD\sps {\gamma_2}(\beta)$ whenever $\gamma_1 \le \gamma_2$.

\subsection{Adversarial Domain Generalization}

We use a relaxation of \eqref{eq:benchmark-error} as our goal of \emph{adversarial domain generalization}, permitting our competitive guarantee to have a small additive slack, parameterized by $\varepsilon$:
\begin{definition}
[$(k,\alpha,\beta,\varepsilon,\delta)$-adversarial domain generalization]
\label[definition]{def:main}
Let $k$ be a positive integer and let $\alpha,\beta,\varepsilon,\delta\in [0,1]$ be parameters satisfying $\beta \ge \alpha$.
We say a learning algorithm $\cL$ solves \emph{$(k,\alpha,\beta,\varepsilon,\delta)$-adversarial domain generalization} for a hypothesis class $\cF$ if the following holds. Let $\cD = (D_1,\ldots,D_k)$ be an arbitrary collection of source domains satisfying the $\alpha$-universality assumption (\Cref{def:alpha-universality}). The learner $\cL$ collects labeled examples from the sources and outputs a classifier $\hat f:\cX\to \{0,1\}$. It may adaptively choose the source of each example as a function of its previous observations and internal randomness. With probability at least $1-\delta$ over the data and the learner's randomness,
\begin{equation}
\label{eq:error-goal}
\err_{D'}(\hat f) \le \max_{f\in \cF_\cD(\beta)}\err_{D'}(f) + \varepsilon \quad \text{for every distribution $D'$ on $\cX\times \{0,1\}$}.
\end{equation}
\end{definition}
We do \emph{not} assume in general that the output $\hat{f}$ belongs to $\cF$;
when we impose $\hat{f}\in\cF$, we say so explicitly. However, if $\hat f$ does belong to the benchmark class $\cF_\cD(\beta)$, which is a subset of $\cF$, then $\hat f$ automatically satisfies \eqref{eq:error-goal}:
\[
\err_{D'}(\hat f) \le \max_{f\in \cF_\cD(\beta)}\err_{D'}(f) \le \max_{f\in \cF_\cD(\beta)}\err_{D'}(f) + \varepsilon.
\]

The error guarantee \eqref{eq:error-goal} is competitive with the worst-case hypotheses in the \emph{strict} benchmark set $\cF_\cD(\beta)$, which consists of hypotheses with error at most $\beta$ on \emph{every} source domain. This guarantee is meaningful when all hypotheses in the strict benchmark set have small error on the target domain. As discussed in the Introduction, we may have reason to believe that hypotheses in the larger, relaxed benchmark set $\cF_\cD\sps \gamma(\beta)$ also achieve low error on the target domain, where the relaxed benchmark set consists of hypotheses that have error at most $\beta$ on at least $1-\gamma$ fraction of the source domains.
In this case, we relax the error guarantee accordingly:
\begin{definition}[$\gamma$-relaxed adversarial domain generalization]
\label[definition]{def:main-relaxed}
    Let $\gamma\in [0,1]$. In the setting of \Cref{def:main}, we say a learner $\cL$ solves \emph{$\gamma$-relaxed} $(k,\alpha,\beta,\varepsilon,\delta)$-adversarial domain generalization for $\cF$ if it achieves the following relaxation of \eqref{eq:error-goal}, where $\cF_\cD(\beta)$ is replaced by $\cF_\cD\sps \gamma(\beta)$:
    \begin{equation}
    \label{eq:error-goal-relaxed}
    \err_{D'}(\hat f) \le \max_{f\in \cF_\cD\sps \gamma(\beta)}\err_{D'}(f) + \varepsilon \quad \text{for every distribution $D'$ on $\cX\times \{0,1\}$}.
    \end{equation}
\end{definition}
Our focus is on the \emph{sample complexity} of adversarial domain generalization. That is, how many data points in total from the $k$ source distributions are needed to solve the adversarial domain generalization task. We establish lower and upper bounds that match up to logarithmic factors in the following sections.

\paragraph{Convention.}
All logarithms in this paper are natural, unless explicitly stated.

\section{Sample Complexity Lower Bounds}
\label[section]{sec:lb}
In this section, we prove sample complexity lower bounds for the adversarial domain generalization problem defined in the previous section.
Our first theorem states the lower bound in terms of the class size.  We convert
it to a VC-dimension bound in \Cref{cor:lb-vc}; the strict case corresponds to
$B=k$ there.

\begin{theorem}[Strict-benchmark lower bound]
\label[theorem]{thm:lb-strict}
    Fix $k\ge1$, $0\le\alpha<\beta<1/2$, $0\le\varepsilon<1/4$, and
    $\delta\in(0,1/4)$. For every integer $M\ge2$, there exists a hypothesis
    class $\cF$ of size $kM+1$ with the following property.
    Let $\cL$ be any learner that solves
    $(k,\alpha,\beta,\varepsilon,\delta)$-adversarial domain generalization
    for $\cF$ in the sense of \Cref{def:main} using at most $n$ data points in total. Then
    \begin{equation}
        \label{eq:strict-lower-bound}
        n \ge \Omega\left(
            k\frac{\beta}{(\beta-\alpha)^2}
            \bigl[\log M+\log(1/\delta)\bigr]
        \right).
    \end{equation}
    Moreover, for each source tuple $\cD=(D_1,\ldots,D_k)$ considered in our lower bound proof, we focus on a \emph{single} target
    distribution $D'$. This distribution is fixed before the source data points
    and the learner's randomness are realized and satisfies a zero-error guarantee for every benchmark classifier:
    \begin{equation}
 \label{eq:realizability}
        \max_{f\in\cF_\cD(\beta)}\err_{D'}(f)=0.
    \end{equation}
    The lower bound continues to hold if the only target-domain guarantee
    required of the learner is
    \begin{equation}
    \label{eq:strict-single-target-goal}
        \Pr\bigl[\err_{D'}(\widehat f)\le\varepsilon\bigr]
        \ge 1-\delta
    \end{equation}
    for this single target domain $D'$ associated with the source tuple $\cD=(D_1,\ldots,D_k)$.
\end{theorem}
Since $\alpha\le\beta$, $\alpha$-universality implies
$f^\star\in\cF_\cD(\beta)$.  Equation~\eqref{eq:realizability} then gives
$\err_{D'}(f^\star)=0$.  Thus the lower bound already holds in the easier setting
where the universal source classifier is also realizable on the target.

Our lower bound for the relaxed benchmark will follow by repeating each domain in the strict
construction sufficiently many times.  The key observation is that repetition preserves $\alpha$-universality. We state the result below and prove it in \Cref{sec:proof-lb}.
\begin{theorem}[Relaxed-benchmark lower bound]
\label[theorem]{thm:lb}
    In the setting of \Cref{thm:lb-strict}, additionally fix a relaxation parameter $\gamma\in [0,1)$ and define
    \[
        r:=\lfloor\gamma k\rfloor+1,
        \qquad
        B:=\lfloor k/r\rfloor.
    \]
    For every integer $M\ge2$, there exists a hypothesis class $\cF$ of size
    $BM+1$ with the following property. Let $\cL$ be any learner that solves $\gamma$-relaxed
    $(k,\alpha,\beta,\varepsilon,\delta)$-adversarial domain generalization
    for $\cF$ in the sense of Definition~\ref{def:main-relaxed} using at most $n$ data points in total. Then
    \begin{equation}
        \label{eq:relaxed-lower-bound}
        n \ge \Omega\left(
            B\frac{\beta}{(\beta-\alpha)^2}
            \bigl[\log M+\log(1/\delta)\bigr]
        \right).
    \end{equation}
Moreover, as in \Cref{thm:lb-strict}, the lower bound continues to hold if the only target-domain guarantee required of the learner is $\Pr\bigl[\err_{D'}(\widehat f)\le\varepsilon\bigr]
        \ge 1-\delta$ for a single target domain $D'$ associated with each source tuple $\cD$ considered in our proof, where $D'$ satisfies $\max_{f\in \cF_\cD\sps \gamma(\beta)}\err_{D'}(f) = 0$.
\end{theorem}
\Cref{cor:lb-vc} gives the VC-dimension versions of
\Cref{thm:lb-strict,thm:lb}.  Setting $\gamma=0$ makes $B=k$ and recovers the
strict case.

\begin{corollary}
    \label[corollary]{cor:lb-vc}
    In the setting of \Cref{thm:lb}, for every positive integer
    $d \ge 10\log B$, there exists a hypothesis class $\cF$ with VC
    dimension at most $d$ such that the total number $n$ of source samples
    used by any learner $\cL$ solving $\gamma$-relaxed
    $(k,\alpha,\beta,\varepsilon,\delta)$-adversarial domain generalization
    for $\cF$ satisfies
    \[
        n\ge \Omega\left(
            B\frac{\beta}{(\beta-\alpha)^2}
            \bigl[d+\log(1/\delta)\bigr]
        \right).
    \]
\end{corollary}
We prove \Cref{cor:lb-vc} using \Cref{thm:lb}. We then prove \Cref{thm:lb-strict,thm:lb} in the next two subsections.
\begin{proof}
    Set
    \[
        M:=\left\lfloor\frac{2^d}{B}\right\rfloor.
    \]
    We first verify that $M\ge2$, as required by \Cref{thm:lb}. If $d=1$,
    then $1\ge10\log B$ forces the positive integer $B$ to equal $1$, and
    hence $M=2$. If $d\ge2$, the assumption $\log B\le d/10$ gives
    \[
        \frac{2^d}{B}
        \ge \exp\bigl((\log 2-1/10)d\bigr)>3,
    \]
    so again $M\ge2$.

    Apply \Cref{thm:lb} with this value of $M$. The resulting class has
    cardinality
    \[
        |\cF|=BM+1\le2^d+1<2^{d+1}.
    \]
    Since a class that shatters $v$ points contains at least $2^v$
    distinct hypotheses, $\operatorname{VC}(\cF)\le\lfloor\log_2|\cF|\rfloor
    \le d$.

    It remains to express the first term of the lower bound in terms of $d$.
    For $x\ge2$, $\lfloor x\rfloor\ge x/2$. Thus, when $d\ge2$,
    \[
        \log M
        \ge d\log2-\log B-\log2
        \ge (\log2-1/10)d-\log2
        \ge \frac d5.
    \]
    The same conclusion holds when $d=1$, because then $B=1$ and $M=2$.
    Substituting $B\log M\ge Bd/5$ into the lower bound of
    \Cref{thm:lb} and absorbing the factor $1/5$ into the universal constant
    proves the claim.
\end{proof}

\subsection{The Hidden Coordinate Lemma}
\label[subsection]{sec:hidden}
Our lower bound proofs %
rely on \Cref{lem:hidden-coordinate,lem:two-coordinate-reference} below, whose full proofs appear in \Cref{app:hidden-coordinate-proof}.
Imagine a learner who observes $M$-bit strings with one hidden exceptional coordinate: the exceptional coordinate has Bernoulli mean $\alpha$, while all other coordinates have Bernoulli mean $\beta$. The learner must identify the exceptional coordinate with high probability. The lemma quantifies the number of samples required for this task.  The heart of the lower bound proof is a reduction from this {\em hidden-coordinate} learning problem to adversarial domain generalization.

Let $M$ be a
positive integer. For every $i\in [M]$, define $D_i(\alpha,\beta)$ to be the
distribution on $M$-bit strings $\{0,1\}^M$ where the (exceptional) $i$-th coordinate is
sampled from the Bernoulli distribution with mean $\alpha$, and all other
coordinates are sampled independently from the Bernoulli distribution with
mean $\beta$.

\begin{lemma}
    \label[lemma]{lem:hidden-coordinate}
    Let $M \ge 2$ be an integer and let
    $0\le\alpha<\beta\le1/2$.
    Let $I$ be uniformly distributed on $[M]$.  A randomized learner
    adaptively observes an
    arbitrary, possibly random, number $N$ of i.i.d.\ examples from
    $D_I(\alpha,\beta)$ and outputs $\widehat I\in[M]$. If
    $\Pr[\widehat I=I]\ge3/4$, then
    \[
        \E[N] \ge \Omega\left(
            \frac{\beta}{(\beta-\alpha)^2}
            \log M
        \right).
    \]
\end{lemma}
The learner is allowed to stop when it feels it has sufficient confidence,
rather than running for some pre-specified number of steps; nonetheless,
Lemma~\ref{lem:hidden-coordinate} shows that the expected number of samples is
large.

The lemma concerns a single $M$-way hidden-coordinate identification task. The $\log M$ term comes from the need to identify one coordinate among $M$ possibilities.  The same
argument shows that this term is already necessary if the learner succeeds with any fixed probability strictly greater than $1/2$.
\begin{remark}
\label[remark]{rem:M-plus-one}
    We will make heavy use of \Cref{lem:hidden-coordinate}.  Later, it will be convenient to think of strings of length $M+1$, with $M\ge 1$ ordinary coordinates and one exceptional coordinate (rather than simply $M \ge 2$ coordinates as in the statement of the lemma).
\end{remark}

We will also need the following similar lemma with a tight dependence on the failure probability bound $\delta$.

\begin{lemma}[Two-coordinate testing under a common reference]
    \label[lemma]{lem:two-coordinate-reference}
    Let $0\le\alpha<\beta\le1/2$ and $\delta\in(0,1/4)$. Define
    \[
        P_1=\operatorname{Bernoulli}(\alpha)
             \otimes\operatorname{Bernoulli}(\beta),
        \qquad
        P_2=\operatorname{Bernoulli}(\beta)
             \otimes\operatorname{Bernoulli}(\alpha),
    \]
    and let
    $R=\operatorname{Bernoulli}(\alpha)^{\otimes2}$.
    A randomized sequential test observes i.i.d.\ samples, stops after an
    arbitrary, possibly random, number $N$ of samples, and outputs
    $\widehat I\in\{1,2\}$. If
    \[
        \Pr_{P_1}[\widehat I=1]\ge1-\delta,
        \qquad
        \Pr_{P_2}[\widehat I=2]\ge1-\delta,
    \]
    then its expected sample size under the reference distribution satisfies
    \[
        \E_R[N]\ge\Omega\left(
            \frac{\beta}{(\beta-\alpha)^2}\log(1/\delta)
        \right).
    \]
\end{lemma}

\noindent

Lemma~\ref{lem:two-coordinate-reference} supplies the separate
confidence-dependent ingredient.  The expected sample size is measured under
the common reference $R=(\alpha,\alpha)$ rather than under either alternative.
This direction of comparison yields the finite divergence
$\kl(\alpha\|\beta)$ even when $\alpha=0$.  Later, we use one common
reference experiment for all source blocks, so their expected query counts
sum to the learner's total budget.

Proofs of both lemmas use standard information theory techniques and appear in Appendix~\ref{app:hidden-coordinate-proof}.

\subsection{Proof of the Strict-Benchmark Lower Bound}
\label[subsection]{sec:proof-lb}
    We prove \Cref{thm:lb-strict} by reducing the hidden-coordinate identification problem in
    Lemma~\ref{lem:hidden-coordinate} to adversarial domain generalization. The two source families used for the size and confidence terms are summarized in \Cref{tab:lower-bound-gadget}. As noted in Remark~\ref{rem:M-plus-one}, it will be easier to think in terms of strings of length $M+1$, with $M\ge 1$ ordinary coordinates and one exceptional coordinate.
With this in mind, we begin by constructing an infinite family of hard instances.  Whereas the hidden-coordinate problem involved $M+1$ distributions, the hard instances here can be viewed as $k \ge 1$ coupled hidden-coordinate problems sharing one distinguished coordinate.

\paragraph{The Hard Instances of Source Domains.}
   Let
    \[
        \eta:=\frac12\min\{\beta-\alpha,1/2-\beta\},
        \qquad \rho:=\beta+\eta.
    \]
    Thus $\beta<\rho<1/2$, while
    \begin{equation}
        \label{eq:rho-comparisons}
        \rho-\alpha\le \frac32(\beta-\alpha),
        \qquad
        \beta\le\rho\le\frac32\beta.
    \end{equation}

    Set $N:=kM+1$, let $\cX:=\{0,1\}^N$, and take the coordinate class
    \[
        \cF:=\{f_1,\ldots,f_N\},
        \qquad f_i(x):=x_i.
    \]
    Thus $|\cF|=N=kM+1$.

\noindent
    A {\em hidden instance} is an ordered partition of $[N]$ into $\{t\}$ together with a set of $k$ equal-sized {\em blocks}:
    \[
        \theta=(t,C_1,\ldots,C_k),
        \qquad
        [N]=\{t\}\sqcup C_1\sqcup\cdots\sqcup C_k,
        \qquad |C_s|=M.
    \]
    The resulting pattern of coordinate-wise error means is displayed in
    \Cref{tab:lower-bound-gadget}(a).

    For this instance and each source index $s\in[k]$, define the source
    distribution $D_s^\theta$ as follows. For every $i\in[N]$, set
    \[
        q_{s,i}^\theta:=
        \begin{cases}
            \rho, & i\in C_s,\\
            \alpha, & \text{otherwise},
        \end{cases}
    \]
    Next draw a fair bit $Y\sim\operatorname{Bernoulli}(1/2)$ and,
    independently, draw
    \[
        \Xi_i\sim\operatorname{Bernoulli}(q_{s,i}^\theta),
        \qquad i\in[N],
    \]
    mutually independently. Set $X_i:=Y\oplus\Xi_i$ for every $i\in[N]$ and
    return $(X,Y)$. Consequently,
    \[
        \mathbf{1}\{f_i(X)\ne Y\}=\Xi_i
        \qquad\text{for every }i\in[N].
    \]
    Thus all $kM+1$ error indicators are mutually independent and
    $\err_{D_s^\theta}(f_i)=q_{s,i}^\theta$ for every $i\in[N]$. In
    particular, $f_t$ has error $\alpha$ on every source, so the sources
    satisfy $\alpha$-universality. Every $f_i$ with $i\in C_s$ has error
    $\rho>\beta$ on source $s$ and is therefore excluded from the strict
    benchmark. Hence
    \begin{equation}
        \label{eq:theorem-one-singleton}
        \cF_{(D_1^\theta,\ldots,D_k^\theta)}(\beta)
        =\{f_t\}.
    \end{equation}
\paragraph{The Target Domain and the Exact Identification Guarantee.}
    Let $D'_t$ be the target distribution obtained by drawing $X$ uniformly
    from $\cX=\{0,1\}^N$ and setting $Y=X_t=f_t(X)$.  By
    \eqref{eq:theorem-one-singleton},
    \[
        \max_{f\in
        \cF_{(D_1^\theta,\ldots,D_k^\theta)}(\beta)}
        \err_{D'_t}(f)
        =\err_{D'_t}(f_t)=0,
    \]
    so $D'_t$ is the single zero-benchmark-error target domain promised in the theorem.
    For binary classifiers $g,h$ on $\cX$, write
    \[
        d(g,h):=\Pr_{X\sim\operatorname{Unif}(\cX)}[g(X)\ne h(X)].
    \]
    If the error guarantee \eqref{eq:error-goal} holds for $D'_t$, then
    \[
        d(\widehat f,f_t)=\err_{D'_t}(\widehat f)\le\varepsilon.
    \]
    For every $j\ne t$, the two distinct coordinate functions $f_j$ and $f_t$ disagree on
    exactly half of the Boolean cube, so $d(f_j,f_t)=1/2$.  The triangle
    inequality and $\varepsilon<1/4$ give
    \[
        d(\widehat f,f_j)
        \ge d(f_j,f_t)-d(\widehat f,f_t)
        \ge \frac12-\varepsilon
        >\varepsilon
        \ge d(\widehat f,f_t).
    \]
    Hence $f_t$ is the unique member of $\cF$ closest to $\widehat f$ under
    $d$, and a successful learner identifies $t$ exactly by nearest-neighbor
    decoding.  The rest of the proof uses only this exact-identification
    consequence, and therefore applies under the weaker single-target
    requirement stated in the theorem.

    This completes the description of the hard instances.

\begin{table}[htbp]
\centering
\begin{minipage}[t]{0.48\textwidth}
\centering
\textbf{(a) Balanced blocks: the \(k\log M\) term}\par\medskip
\begin{tabular}{c|ccccc}
\toprule
source & \(t\) & \(C_1\) & \(C_2\) & \(\cdots\) & \(C_k\)\\
\midrule
\(D_1^\theta\) & \(\alpha\) & \(\rho\) & \(\alpha\) & \(\cdots\) & \(\alpha\)\\
\(D_2^\theta\) & \(\alpha\) & \(\alpha\) & \(\rho\) & \(\cdots\) & \(\alpha\)\\
\(\vdots\)     & \(\vdots\)& \(\vdots\)& \(\vdots\)& \(\ddots\)& \(\vdots\)\\
\(D_k^\theta\) & \(\alpha\) & \(\alpha\) & \(\alpha\) & \(\cdots\) & \(\rho\)\\
\bottomrule
\end{tabular}
\par\medskip
\(\lvert C_s\rvert=M\), while the shared exceptional coordinate \(t\)
has error \(\alpha\) on every source.
\end{minipage}
\hfill
\begin{minipage}[t]{0.48\textwidth}
\centering
\textbf{(b) Hidden extra coordinate: the \(k\log(1/\delta)\) term}\par\medskip
\begin{tabular}{c|cc}
\toprule
source \(s\) & \(u\) & \(v\)\\
\midrule
reference \(R_s\) & \(\alpha\) & \(\alpha\)\\
\(D_s^{u,s}\)     & \(\alpha\) & \(\rho\)\\
\(D_s^{v,s}\)     & \(\rho\)   & \(\alpha\)\\
\bottomrule
\end{tabular}
\par\medskip
All other observations have the same distribution. Under the reference
transcript law \(Q^\Pi\), the query counts satisfy
\(\sum_s\E_{Q^\Pi}[N_s]\le n\).
\end{minipage}
\caption{\textbf{The two source families in the proof of
\Cref{thm:lb-strict}.}
In part (a), the coordinates are partitioned as
\([N]=\{t\}\sqcup C_1\sqcup\cdots\sqcup C_k\). Source \(s\) assigns error
mean \(\rho\) to the \(M\) coordinates in \(C_s\) and error mean \(\alpha\)
to every other coordinate. Conditioning on all but \(C_s\cup\{t\}\)
produces an \((M+1)\)-way hidden-coordinate problem for each source, giving
the \(k\log M\) term. Part (b) shows the second family for a fixed possible
owner \(s\): the alternatives interchange which of \(u,v\) is exceptional,
whereas the common reference keeps both at mean \(\alpha\). Applying the
two-coordinate reference lemma to every \(s\) and summing the reference
query counts gives the \(k\log(1/\delta)\) term.}
\label{tab:lower-bound-gadget}
\end{table}

\paragraph{The Blockwise Bound.}
This is the principal reduction from hidden-coordinate identification to
adversarial domain generalization; see \Cref{tab:lower-bound-gadget}(a).
Given an instance of the hidden-coordinate problem, we choose a random distinguished set $U_s$ of size $M+1$ indices in $[N]$ and $k-1$ other random source blocks of $M$ indices each. Randomly biject the coordinates of the hidden-coordinate instance to $U_s$. We map the hidden exceptional coordinate to $t$, and define $C_s$ to be the other $M$ coordinates of $U_s$.  This creates one coupled identification problem for each source block.

Conditioning on all information outside the coordinates $C_s\cup\{t\}$ yields an instance of the hidden coordinate problem,
so \Cref{lem:hidden-coordinate} lower-bounds the conditional
expected number of queries $N_s$ to that source and hence $\E[N_s]$.  Because
$\sum_{s=1}^kN_s\le n$, the $k$ per-source bounds add to give the factor $k$.

Choose a hidden partition $\theta=(t,C_1,\dots,C_k)$ uniformly from the set of all hidden partitions on $[N]$.
    For each
    $s\in[k]$, let $N_s$ be the random number of rounds on which $\cL$ queries
    source $s$. Thus
    \begin{equation}
        \label{eq:adaptive-query-counts}
        \sum_{s=1}^k N_s\le n
        \qquad\text{on every transcript.}
    \end{equation}
    Recall that $D_t(\alpha,\rho)$ is the hidden-coordinate distribution in which coordinate $t$ is the special low-noise coordinate.
    Fix $s$, condition on the unordered set $U_s:=\{t\}\cup C_s$, and
    condition on all other blocks $C_{s'}$, $s'\ne s$. Under this conditioning, everything about the partition is known except which member of $U_s$ is the special coordinate $t$, which
    is uniform on $U_s$.
On a query to source $s$, the learner receives a labeled example $(X,Y)$, where $X\in\{0,1\}^N$ and $Y\in\{0,1\}$. Since $f_i(X)=X_i$, the learner can evaluate the hypotheses $f_i$ for each $i\in U_s$ and form the vector of error indicators
$$\left(\mathbf 1\{f_i(X)\neq Y\}\right)_{i\in U_s}.$$
By the construction of $D_s^\theta$, this vector consists of independent Bernoulli variables, with mean $\alpha$ at the distinguished coordinate $t$ and mean $\rho$ at every coordinate in $C_s$. Thus, after relabeling the coordinates in $U_s$ as $[M+1]$, each query to source $s$ produces one sample from the $(M+1)$-way hidden-coordinate distribution $D_t(\alpha,\rho)$.
    For a query to any other source, this vector has a probability distribution
    independent of $t$. The remaining error indicators also have a distribution
    independent of $t$.
    Moreover\, given the vector of error indicators $(\Xi_i)_{i\in[N]}$ with the specified distribution, one can generate a labeled sample distributed according to $D_s^\theta$ by drawing an independent fair bit $Y$ and setting $X_i=Y\oplus\Xi_i$ for every $i\in[N]$.
    Hence the fair label carries no additional information about $t$.

    Consequently, running $\cL$ gives a hidden-coordinate learner that uses
    $N_s$ oracle samples and identifies the uniform index $t\in U_s$ with
    probability at least $1-\delta$. Lemma~\ref{lem:hidden-coordinate},
    with $\rho$ in place of $\beta$, therefore gives
    \[
        \E[N_s\mid U_s,(C_{s'})_{s'\ne s}]
        \ge \Omega\left(
            \frac{\rho}{(\rho-\alpha)^2}
            \log|U_s|
        \right).
    \]
    Since $|U_s|=M+1$, averaging the conditioning and then summing over
    $s$ in \eqref{eq:adaptive-query-counts} yields
    \begin{equation}
        \label{eq:theorem-one-size-bound}
        n\ge \Omega\left(
            k\frac{\rho}{(\rho-\alpha)^2}
            \log M
        \right).
    \end{equation}
    This argument permits the source queried at every round to depend
    arbitrarily on the preceding transcript.

\paragraph{The Confidence-Dependent Bound.}
It remains to prove the factor-$k$ confidence term. We use a second family
of source tuples on the same coordinate class, summarized in
\Cref{tab:lower-bound-gadget}(b). Fix distinct coordinates
$u,v\in[N]$ and an owner map
\[
    g:[N]\setminus\{u,v\}\longrightarrow[k]
\]
having one fiber of size $M-1$ and the other $k-1$ fibers of size $M$.
For every $s\in[k]$, form two source tuples. In $\cD^{u,s}$, coordinate
$u$ is exceptional, the coordinates governed by $g$ retain their owners,
and $v$ is assigned to source $s$. In $\cD^{v,s}$, interchange the roles of
$u$ and $v$. Thus the owner blocks have sizes
\[
    |C_j|=|g^{-1}(j)|+\mathbf 1\{j=s\},
\]
and the identity of the block receiving the extra coordinate is not public.
Define the source distributions from these blocks exactly as above, using
error mean $\rho$ on the owner block and $\alpha$ elsewhere. Consequently,
\[
    \cF_{\cD^{u,s}}(\beta)=\{f_u\},
    \qquad
    \cF_{\cD^{v,s}}(\beta)=\{f_v\}.
\]
Associate these tuples with the targets $D_u'$ and $D_v'$, respectively,
where $X$ is uniform on $\{0,1\}^N$ and $Y=X_u$ or $Y=X_v$. The
nearest-coordinate decoder above therefore produces an estimate
$\widehat t$ satisfying, for every $s$,
\begin{equation}
    \Pr_{\cD^{u,s}}[\widehat t=u]\ge1-\delta,
    \qquad
    \Pr_{\cD^{v,s}}[\widehat t=v]\ge1-\delta.
    \label{eq:strict-hidden-block-success}
\end{equation}

Define one common reference source tuple
    $\mathcal R=(R_1,\ldots,R_k)$. Under $R_j$, the coordinates in
    $g^{-1}(j)$ have error mean $\rho$, while every other coordinate,
    including $u$ and $v$, has error mean $\alpha$. Let $\Pi$ denote the
    learner's complete transcript and let $Q^\Pi$ be its distribution under
    $\mathcal R$. As above, $N_s$ is the number of queries to source $s$.
    Since the learner uses at most $n$
    samples,
    \begin{equation}
        \sum_{s=1}^k\E_{Q^\Pi}[N_s]\le n.
        \label{eq:strict-reference-query-sum}
    \end{equation}

    Fix $s$. Using a sample from
    $\operatorname{Bernoulli}(\alpha)\otimes
      \operatorname{Bernoulli}(\rho)$ (or, respectively, from
    $\operatorname{Bernoulli}(\rho)\otimes
      \operatorname{Bernoulli}(\alpha)$) whenever the learner queries source $s$, and generating all other observations locally, simulates
    the interactions with $\cD^{u,s}$ (or, respectively, $\cD^{v,s}$). Under
    the two-coordinate reference
    $\operatorname{Bernoulli}(\alpha)^{\otimes2}$, the same simulation has
    transcript distribution $Q^\Pi$ and uses exactly $N_s$ oracle samples.
    Therefore \eqref{eq:strict-hidden-block-success} and
    Lemma~\ref{lem:two-coordinate-reference}, with $\rho$ in place of
    $\beta$, imply for every $s$ that
    \[
        \E_{Q^\Pi}[N_s]\ge\Omega\left(
            \frac{\rho}{(\rho-\alpha)^2}\log(1/\delta)
        \right).
    \]
    Summing over $s$ and using
    \eqref{eq:strict-reference-query-sum} yields the complete
    confidence-dependent bound
    \begin{equation}
        n\ge\Omega\left(
            k\frac{\rho}{(\rho-\alpha)^2}\log(1/\delta)
        \right).
        \label{eq:strict-reference-confidence}
    \end{equation}

\paragraph{Completing the Proof.} We are now ready to complete the proof of Theorem~\ref{thm:lb-strict}.
\begin{proof}[Proof of Theorem~\ref{thm:lb-strict}]
    Retaining only the $\log M$ term in the balanced-block bound
    \eqref{eq:theorem-one-size-bound} gives
    \[
        n\ge\Omega\left(
            k\frac{\rho}{(\rho-\alpha)^2}\log M
        \right).
    \]
    The comparison \eqref{eq:rho-comparisons} gives
    \[
        \frac{\rho}{(\rho-\alpha)^2}
        \ge\frac49\frac{\beta}{(\beta-\alpha)^2}.
    \]
    Applying this comparison both to the preceding size bound and to
    \eqref{eq:strict-reference-confidence}, and then using
    $\max\{a,b\}\ge(a+b)/2$, proves
    \eqref{eq:strict-lower-bound}.
\end{proof}

\subsection{Proof of the Relaxed-Benchmark Lower Bound}
\label[subsection]{sec:proof-lb-relaxed}
We prove \Cref{thm:lb} by repeating each source domain from the strict
construction sufficiently often.
\begin{proof}[Proof of Theorem~\ref{thm:lb}]
    Let
    \[
        r=\lfloor\gamma k\rfloor+1,
        \qquad
        B=\lfloor k/r\rfloor,
    \]
    as in the theorem. Instantiate Theorem~\ref{thm:lb-strict} with $B$
    source domains, and let $\cF$ be the resulting hypothesis class of size
    $BM+1$.

    Partition the $k$ source indices into
    \[
        G_b:=\{(b-1)r+1,\ldots,br\}\quad (b<B),
        \qquad
        G_B:=\{(B-1)r+1,\ldots,k\}.
    \]
    Each group has size at least $r$. Given any $B$-tuple
    $\widetilde{\cD}=(\widetilde D_1,\ldots,\widetilde D_B)$, form a
    $k$-tuple $\cD=(D_1,\ldots,D_k)$ by setting
    \[
        D_s:=\widetilde D_b
        \qquad\text{whenever }s\in G_b.
    \]
    Repetition preserves $\alpha$-universality. It also gives the exact
    benchmark identity
    \begin{equation}
        \label{eq:repetition-benchmark-identity}
        \cF_{\cD}^{(\gamma)}(\beta)
        =\cF_{\widetilde{\cD}}(\beta).
    \end{equation}
    Indeed, a hypothesis in the strict benchmark on the right violates no
    threshold and therefore belongs to the relaxed benchmark on the left.
    Conversely, if a hypothesis is not in the strict benchmark, then it has
    error greater than $\beta$ on some $\widetilde D_b$. It consequently
    violates the threshold on every source in $G_b$, hence on at least
    $r>\gamma k$ sources, and is excluded from the relaxed benchmark.

    Now suppose $\cL$ is a learner satisfying the premise of
    Theorem~\ref{thm:lb}. Construct a learner $\widetilde{\cL}$ for $B$ source
    domains by running $\cL$ on the repeated tuple above. Whenever $\cL$
    queries a source $s\in G_b$, the learner $\widetilde{\cL}$ requests a
    fresh sample from $\widetilde D_b$ and returns it to $\cL$. This simulates
    the repeated-domain interaction exactly, including when the next source is
    selected adaptively, and uses the same total budget of at most $n$ samples.

    By \eqref{eq:repetition-benchmark-identity}, the relaxed guarantee of
    $\cL$ becomes the strict guarantee of $\widetilde{\cL}$. The same identity
    shows that the restricted-target version of the premise also becomes the
    restricted-target version in Theorem~\ref{thm:lb-strict}. That theorem,
    with $B$ in place of $k$, therefore gives
    \[
        n \ge \Omega\left(
            B\frac{\beta}{(\beta-\alpha)^2}
            \bigl[\log M+\log(1/\delta)\bigr]
        \right),
    \]
    as claimed.
\end{proof}

\section{Sample Complexity Upper Bounds}
\label[section]{sec:ub}

In this section, we prove sample complexity upper bounds that match the lower bounds in
\Cref{sec:lb} up to logarithmic factors.
We state the tightest upper bound in \Cref{thm:ub} as well as a simpler yet slightly weaker bound in \Cref{cor:ub-uniform}. When the relaxation parameter $\gamma$ is a constant, we obtain a simple upper bound in \Cref{cor:ub-constant-gamma}.

To solve the adversarial domain generalization problem, it is enough to find any hypothesis in the strict or relaxed benchmark set $\cF_\cD(\beta)$ or $\cF_\cD\sps \gamma(\beta)$:
such a proper output satisfies the competitive error guarantee by definition.  Our algorithm therefore aims at rejecting every hypothesis whose error exceeds $\beta$ on
at least $r:=\lfloor\gamma k\rfloor+1$ sources without rejecting the
$\alpha$-universal hypothesis.  There are two useful ways to aggregate this
evidence, corresponding to two algorithms for different parameter regimes.
We give some high level intuition here.  Details appear in \Cref{subsec:ub-algorithms}.

The subset-certificate learner tests every subset of $r$ source domains. A function $f$ that exceeds error $\beta$ on $r$ or more sources has at least one such subset on which its average error exceeds $\beta$, and is therefore rejected by the corresponding empirical test with high probability. The $\alpha$-universal hypothesis has average error at most $\alpha$ on every subset and passes all the tests with high probability.

The {\em random-domain batch learner} repeatedly chooses a source domain at random and draws a small batch of examples from that domain. It uses each batch to determine whether a candidate function’s empirical error exceeds a threshold $q=(\alpha+\beta)/2$. The $\alpha$-universal hypothesis exceeds this threshold only rarely, whereas any function outside the relaxed benchmark exceeds it on at least $r$ sources (by definition), and therefore does so with noticeably larger probability when the source is chosen at random. Repeating the experiment sufficiently many times lets the learner distinguish these two cases with high probability.

Throughout this section, assume $\gamma\in[0,1/4]$ and write, as in the
relaxed lower bound,
\[
    r:=\lfloor \gamma k\rfloor+1,\qquad
    B:=\left\lfloor\frac{k}{r}\right\rfloor,\qquad
    \kappa:=\frac{\beta}{(\beta-\alpha)^2},\qquad
    \ell:=\log(eB).
\]
Since $B\le k/r<2B$, replacing $k/r$ by $B$ affects our bounds only by a
universal constant.  Moreover, for $k\ge2$ our assumption on $\gamma$ gives
$r\le k/2$, and the standard binomial estimates yield
\[
    \log\binom{k}{r}
    =\Theta\!\left(r\log\frac{ek}{r}\right)
    =\Theta(r\ell).
\]
When $k=1$, the binomial term is zero and is absorbed by the other terms in
the bounds below.  We also define the separation parameter
\[
    \rho_{\alpha,\beta}
    :=
    \begin{cases}
        1+\log(\beta/\alpha), & \alpha>0,\\
        +\infty, & \alpha=0,
    \end{cases}
    \qquad
    A_{\alpha,\beta}(k,r)
    :=1+\frac{\ell}{\rho_{\alpha,\beta}},
\]
where a quotient by \(+\infty\) is zero.

\begin{theorem}[Best-of-two upper bound]
\label[theorem]{thm:ub}
Let \(k\ge1\), \(0\le \alpha<\beta<1/2\), \(\gamma\in[0,1/4]\),
\(\varepsilon\in[0,1]\), \(\delta\in(0,1)\), and let
\(\cF\) have VC dimension at most an integer \(d\ge1\).  There is a
nonadaptive proper learner that achieves \(\gamma\)-relaxed
\((k,\alpha,\beta,\varepsilon,\delta)\)-adversarial domain
generalization using
\begin{equation}
\label{eq:upper-bound-main}
    n=O\!\left(
        \kappa B\min\left\{
        \begin{array}{l}
            d\log(e\kappa)+\log(1/\delta)+r\ell,
            \\[1mm]
            A_{\alpha,\beta}(k,r)
            \left[
                d\log\!\left(
                    e\kappa B A_{\alpha,\beta}(k,r)
                \right)
                +\log(1/\delta)
            \right]
        \end{array}
        \right\}
    \right)
\end{equation}
labeled source examples.
\end{theorem}

The next corollary suppresses the finer comparison between the two rejection
strategies and is often the easiest form to compare with the lower bound.

\begin{corollary}[Simplified upper bound]
\label[corollary]{cor:ub-uniform}
Under the assumptions of \Cref{thm:ub},
\begin{equation}
\label{eq:upper-bound-simple}
    n
    =
    O\!\left(
        \kappa B\ell
        \left[
            d\log(e\kappa B\ell)+\log(1/\delta)
        \right]
    \right).
\end{equation}
\end{corollary}

When a constant fraction of source domains may be discarded, the
dependence on the total number of domains disappears.

\begin{corollary}[Constant relaxation level]
\label[corollary]{cor:ub-constant-gamma}
For every constant \(\gamma_0\in(0,1/4]\), if
\(\gamma\in[\gamma_0,1/4]\), then
\begin{equation}
\label{eq:upper-bound-constant-gamma}
    n
    =
    O_{\gamma_0}\!\left(
        \kappa
        \left[
            d\log(e\kappa)+\log(1/\delta)
        \right]
    \right).
\end{equation}
In particular, whenever \(\gamma\) is bounded below by a positive constant,
the upper bound is independent of \(k\).
\end{corollary}

In the regime of \Cref{cor:lb-vc}, these results
match the lower bound up to logarithmic factors: the common
leading term is
$\frac{\beta}{(\beta-\alpha)^2}B[d+\log(1/\delta)]$.  In particular,
$B=k$ for the strict benchmark $\gamma=0$, whereas
$B=\Theta(1/\gamma)$ when $\gamma\ge 1/k$.

\subsection{Algorithms for identifying a benchmark hypothesis}
\label[subsection]{subsec:ub-algorithms}
We describe the two algorithms we use to establish \Cref{thm:ub}.
Both algorithms are proper empirical procedures.  As usual for
VC-type sample-complexity statements, we suppress computational issues
associated with carrying out empirical minimization over
\(\cF\).

\paragraph{Subset-certificate learner.}
Roughly speaking, the subset-certificate learner tests every subset of $r$ source domains. The intuition is that if $f$ is not in the relaxed benchmark, then its error exceeds $\beta$ on every domain in some subset of $r$ source domains; hence its error on the uniform mixture of those domains also exceeds $\beta$. In contrast, the $\alpha$-universal hypothesis has error at most $\alpha$ on every source mixture
and therefore survives all of the subset-based rejection tests with high probability.

In more detail, set \(q=(\alpha+\beta)/2\), and let
\(\mathcal S_r=\{S\subseteq[k]:|S|=r\}\).  Draw source indices
independently and uniformly from \([k]\), and then draw one labeled
example from the selected source.  For every \(S\in\mathcal S_r\), use
the first
\[
    m_{\mathrm{cert}}
    =
    \left\lceil
        C\kappa
        \left[
            d\log(e\kappa)
            +\log(1/\delta)
            +r\ell
        \right]
    \right\rceil
\]
observations whose source index lies in \(S\), where $C>0$ is a sufficiently large universal constant arising from the VC threshold-separation bound.  These observations are
i.i.d.\ from the uniform mixture
\(D_S=r^{-1}\sum_{i\in S}D_i\).  The learner returns any
\(f\in\cF\) whose empirical error on every one of these
\(\binom{k}{r}\) samples is at most \(q\).  A random uniform source schedule of
length \(O(Bm_{\mathrm{cert}})\) supplies all of the required samples
with high probability.  Uniform convergence shows that an
\(\alpha\)-universal hypothesis survives all of the subset-based rejection tests, whereas each
hypothesis outside the relaxed benchmark fails the certificate formed
by any \(r\) of its bad domains.  This gives the
first expression in
\eqref{eq:upper-bound-main}.

\paragraph{Random-domain batch learner.}
At a high level, the random-domain batch learner operates in rounds.  At each round, a random domain is selected together with a number $m_0$ of samples from that domain.
Every time a batch is drawn, we evaluate the empirical loss of each $f \in \cF$ on this batch, and record whether or not it exceeds a specific threshold $q$.  Let the {\em score} of a function $f \in \cF$ be the total number of times its loss exceeds $q$ as this process is over a specific number $T$ of rounds.  We return any function with a minimal score.

In more detail, set
\[
    m_0=\left\lceil C\kappa
        A_{\alpha,\beta}(k,r)\right\rceil.
\]
A batch is generated by choosing \(I\) uniformly from \([k]\) and
drawing \(m_0\) independent labeled examples from \(D_I\).  For
\(f\in\cF\), declare the batch loss to be one when the empirical
error of \(f\) on the batch exceeds \(q\), and zero otherwise.  Draw
\[
    T
    =
    O\!\left(
        B\left[
            d\log(e m_0 B)+\log(1/\delta)
        \right]
    \right)
\]
independent batches and return a hypothesis minimizing its empirical
batch loss.  The choice of \(m_0\) makes the batch-loss probability at
most \(r/(32k)\) for an \(\alpha\)-universal hypothesis and larger than
\(3r/(4k)\) for every hypothesis outside the relaxed benchmark.  A
relative VC bound for the induced batch-loss class therefore separates
the two sets using \(T\) batches.  This yields the
second expression in \eqref{eq:upper-bound-main}.

The learner in \Cref{thm:ub} evaluates the two displayed budgets from
the known parameters and runs the cheaper construction.  Complete
proofs, including the simultaneous VC arguments, appear in
\Cref{app:upper-bound-proofs}.

\section{Lower Bounds with Target Access}
\label[section]{sec:target-samples}

It is natural to ask whether even a modest labeled sample from the target
domain can substantially reduce the amount of source data needed for
adversarial domain generalization. \Cref{thm:m-target-lower} shows that, in
general, it cannot:
even after giving the learner $m$ labeled target examples, the learner still needs essentially the same number of source samples as in the no-target-access lower bound—unless $m$ is large enough for the learner to learn the relevant target hypothesis directly from the target data. To be specific, the lower bounds on the number of source samples persist unless the target sample
is on the order of
$(\log M+\log(1/\delta))/\varepsilon^2$. This is essentially the usual
target-only sample size for learning the relevant finite hypothesis class
to excess error $\varepsilon$. Thus target data become useful in the worst
case only when there are enough of them to learn directly from the target,
without relying on the source domains.

We first define the learning task obtained by augmenting
\Cref{def:main} with labeled target data.

\begin{definition}[$m$-target-access adversarial domain generalization]
    \label[definition]{def:target-access}
    Let $k$ be a positive integer, let $m$ be a nonnegative integer, and let
    $\alpha,\beta,\varepsilon,\delta\in[0,1]$ satisfy $\beta\ge\alpha$.
    We say that a learner $\cL$ solves
    \emph{$(k,\alpha,\beta,\varepsilon,\delta)$-adversarial domain
    generalization with $m$-sample target access} for a hypothesis class
    $\cF$ if the following holds. Let
    $\cD=(D_1,\ldots,D_k)$ be an arbitrary tuple of source domains satisfying
    $\alpha$-universality, and let $D'$ be an arbitrary target distribution.
    Before querying the source domains, $\cL$ receives
    $m$ i.i.d.\ labeled examples from $D'$. It may then choose each source
    query adaptively as a function of the target sample, its preceding source
    observations, and its internal randomness, and outputs a classifier
    $\widehat f:\cX\to\{0,1\}$. With probability at least $1-\delta$ over the
    target sample, the source samples, and the learner's randomness,
    \begin{equation}
        \label{eq:target-access-goal}
        \err_{D'}(\widehat f)
        \le
        \max_{f\in\cF_\cD(\beta)}\err_{D'}(f)+\varepsilon.
    \end{equation}
    The $m$ target examples are not included in the source-sample complexity.
\end{definition}

Thus, for every fixed pair $(\cD,D')$, the success probability in
\eqref{eq:target-access-goal} is at least $1-\delta$; unlike in
\Cref{def:main}, the output may depend on labeled data from the
particular target $D'$. The following result shows that, for a noisy target,
the source lower bound persists until the target sample size reaches the
$1/\varepsilon^2$ information scale.

\begin{theorem}[Noisy target lower bound at the $1/\varepsilon^2$ scale]
    \label[theorem]{thm:m-target-lower}
    Fix $k\ge1$, $0\le\alpha<\beta<1/2$,
    $0<\varepsilon<1/4$, $\delta\in(0,1/4)$, and integers $m\ge0$ and
    $M\ge2$. There is a universal constant $c_0>0$, a hypothesis class
    $\cF$ of size $kM+1$, and a family of $\alpha$-universal source tuples
    $\cD^\theta=(D_1^\theta,\ldots,D_k^\theta)$ with associated target
    distributions $D_\theta'$ and numbers $p_\theta\in[0,1/2)$ such that
    \[
        \max_{f\in\cF_{\cD^\theta}(\beta)}\err_{D_\theta'}(f)
        =p_\theta.
    \]
    Suppose a learner receives $m$ i.i.d.\ labeled examples from $D_\theta'$, then
    uses a fixed total budget of $n$ adaptively selected source samples, and
    satisfies, for every hidden instance $\theta$,
    \[
        \Pr\bigl[
            \err_{D_\theta'}(\widehat f)\le p_\theta+\varepsilon
        \bigr]\ge1-\delta.
    \]
    Then
    \begin{equation}
        \label{eq:target-access-epsilon-squared-lower}
        \max\left\{
            \frac{(\beta-\alpha)^2}{\beta}\,n,
            \;k\varepsilon^2m
        \right\}
        \ge c_0 k\bigl[\log M+\log(1/\delta)\bigr].
    \end{equation}
\end{theorem}

The realizable counterpart, \Cref{thm:target-access-epsilon}, strengthens the
target condition by requiring the benchmark error on the target to be zero.  Thus, any candidate function with target error at least $\varepsilon$ is contradicted by a fresh labeled target example with probability at least $\varepsilon$.  So in the realizable case the target contribution is $k\varepsilon m$ rather than
$k\varepsilon^2m$, so the source lower bound persists until the target sample
reaches order $(\log M+\log(1/\delta))/\varepsilon$. This change is
unavoidable: \Cref{prop:zero-target-upper} shows that this many target samples
suffice, up to constants and the distinction between $\log M$ and
$\log|\cF|$, for target-only learning.

The next corollary states the consequence in the VC-dimension language used
in the Introduction.  It covers the strict benchmark by setting $\gamma=0$.

\begin{corollary}[VC-dimension target-access tradeoff]
    \label[corollary]{cor:target-access-vc}
    Fix $k\ge1$, $\gamma\in[0,1)$, $0\le\alpha<\beta<1/2$,
    $0<\varepsilon<1/4$, $\delta\in(0,1/4)$, and a nonnegative integer
    $m$.  Define
    \[
        r:=\lfloor\gamma k\rfloor+1,
        \qquad B:=\lfloor k/r\rfloor.
    \]
    For every positive integer $d\ge10\log B$, there is a hypothesis class
    $\cF$ of VC dimension at most $d$ for which every learner that receives
    $m$ labeled target examples, uses $n$ labeled source examples, and
    satisfies \eqref{eq:target-access-goal} with
    $\cF_\cD^{(\gamma)}(\beta)$ in place of $\cF_\cD(\beta)$ must obey
    \[
        \max\left\{
            \frac{(\beta-\alpha)^2}{\beta}\,n,
            \;B\varepsilon^2m
        \right\}
        =\Omega\!\left(B[d+\log(1/\delta)]\right).
    \]
\end{corollary}

Thus, unless the number of target examples is
$\Omega((d+\log(1/\delta))/\varepsilon^2)$, the source lower bound persists.
We prove \Cref{thm:m-target-lower,cor:target-access-vc} in
\Cref{app:epsilon-target-access}, which also contains the realizable
counterpart \Cref{thm:target-access-epsilon}, the target-only upper bound
\Cref{prop:zero-target-upper}, and the relaxed-benchmark reduction
\Cref{cor:relaxed-target-access}.

\section{Relaxed Benchmarks and Meta-Distribution Generalization}
\label[section]{sec:mdl-generalization}

\Cref{thm:ub} gives a proper learner whose output lies in the relaxed
benchmark class and is therefore competitive with the worst function in that
class on every target domain.  The goal of this section is to give a guarantee
for the worst function in the benchmark when the observed domains are drawn
i.i.d.\ from a meta-distribution.  Intuitively, if even the worst benchmark
function performs well on most of the observed random domains, then it should
also perform well on a new domain drawn from the same meta-distribution.  In
\Cref{subsec:meta-benchmark}, we formalize this intuition using the domain
shattering dimension of Dwork, Hu, and Shao~\citep{dhs} and prove
\Cref{thm:relaxed-transfer}.  In
\Cref{subsec:meta-sample-complexity}, we combine this guarantee with our
earlier upper bound and explain the resulting sample complexity improvement
over Dwork et al.

\subsection{A guarantee for the worst benchmark function}
\label[subsection]{subsec:meta-benchmark}

Let $\cG\subseteq\Delta(\cX\times\{0,1\})$ be a family of domains,
let $P$ be a meta-distribution over $\cG$, and draw
$\cD=(D_1,\ldots,D_k)\sim P^k$.  For a predictor
$f:\cX\to\{0,1\}$, recall that
\[
    \err_D(f)=\Pr_{(x,y)\sim D}[f(x)\neq y].
\]
Following Dwork et al.~\cite[Section~3]{dhs}, for a threshold $t\in[0,1]$
define the \emph{domain-error rate}
\[
    \operatorname{Er}_{P,t}(f)
    :=\Pr_{D\sim P}[\err_D(f)>t].
\]

For a threshold $\tau$ and relaxation parameter $\gamma\in[0,1]$, the
relaxed benchmark from \Cref{def:relaxed-benchmark} is
\[
    \cF_\cD^{(\gamma)}(\tau)
    :=\left\{f\in\cF:
        \frac1k\sum_{i=1}^k
        \mathbf 1[\err_{D_i}(f)>\tau]\le\gamma
      \right\}.
\]

We now recall the domain shattering dimension using the conventions of
Dwork et al.~\cite[Definition~4.1]{dhs}.  For a level $t\in[0,1]$ and a
margin $\rho>0$, a finite set $S\subseteq\cG$ is \emph{$\rho$-shattered by
$\cF$ at level $t$} if, for every $E\subseteq S$, there is an $f_E\in\cF$
such that
\[
    \err_D(f_E)<t-\rho \quad(D\in E),
    \qquad
    \err_D(f_E)>t \quad(D\in S\setminus E).
\]
The domain shattering dimension $\Gdim(\cF,\cG,t,\rho)$ is the largest
cardinality of such a set.

\begin{theorem}[Relaxed benchmark generalizes over domains]
\label[theorem]{thm:relaxed-transfer}
Let $\cF\subseteq\{0,1\}^{\cX}$, let
$\cG\subseteq\Delta(\cX\times\{0,1\})$, and let $P$ be any
meta-distribution over $\cG$.  Draw
$\cD=(D_1,\ldots,D_k)\sim P^k$.  Fix $\tau\in[0,1]$, $\eta>0$ with
$\tau+\eta\le1$, $\gamma\in[0,1]$, and $\delta\in(0,1)$, and let
\[
    d_{\mathrm{dom}}
    :=\Gdim(\cF,\cG,\tau+\eta,\eta/2),
    \qquad
    a_{\mathrm{dom}}
    :=\frac{d_{\mathrm{dom}}\log^2(2k)+\log(2/\delta)}{k}.
\]
There is a universal constant $C$ such that, with probability at least
$1-\delta$ over $\cD$, every $f\in\cF_\cD^{(\gamma)}(\tau)$ satisfies
\[
    \operatorname{Er}_{P,\tau+\eta}(f)
    \le\gamma+C\left(
        \sqrt{\gamma a_{\mathrm{dom}}}+a_{\mathrm{dom}}
    \right).
\]
Consequently, whenever the relaxed benchmark is nonempty,
\[
    \sup_{f\in\cF_\cD^{(\gamma)}(\tau)}
    \operatorname{Er}_{P,\tau+\eta}(f)
    \le\gamma+C\left(
        \sqrt{\gamma a_{\mathrm{dom}}}+a_{\mathrm{dom}}
    \right).
\]
\end{theorem}

The proof uses the following extension of the partial-concept
uniform convergence lemma of Dwork et al.~\cite[Lemma~4.2]{dhs}.  Recall that
a set is shattered by a partial concept class if every binary labeling of the
set is realized without using $\bot$; its VC dimension is the largest size of
such a set~\cite[Definition~2.1]{dhs}.
\begin{lemma}[Relative one-sided uniform convergence for partial concepts]
\label[lemma]{lem:relative-partial}
Let $\cC$ be a class of partial concepts
$c:\mathcal Z\to\{0,1,\bot\}$ with VC dimension at most $d$, and let
$Z_1,\ldots,Z_k$ be drawn i.i.d.\ from a distribution $Q$ on $\mathcal Z$.
There is a universal constant $C$ such that, writing
\[
    a:=\frac{d\log^2(2k)+\log(2/\delta)}{k},
\]
we have
\[
    \Pr_{(Z_1,\ldots,Z_k)\sim Q^k}\!\left[
        \begin{aligned}
        \exists c\in\cC:\quad&
        \left\{
            \frac1k\sum_{i=1}^k\mathbf 1[c(Z_i)\ne0]\le\gamma
        \right\}
        \\
        &{}\text{and}\quad
        \left\{
            \Pr_{Z\sim Q}[c(Z)=1]
            >\gamma+C\left(\sqrt{\gamma a}+a\right)
        \right\}
        \end{aligned}
    \right]
    \le\delta.
\]
\end{lemma}

\begin{proof}
For a sample $S=(Z_1,\ldots,Z_k)$ and a partial concept $c$, write
\[
    \widehat q_S(c)
    :=\frac1k\sum_{i=1}^k\mathbf 1[c(Z_i)\ne0],
    \qquad
    p(c):=\Pr_{Z\sim Q}[c(Z)=1].
\]
Set $s=C(\sqrt{\gamma a}+a)$, where the universal constant $C\ge1$ will be
chosen below, and define the bad event
\[
    A:=\{\exists c\in\cC:\ \widehat q_S(c)\le\gamma
         \text{ and }p(c)>\gamma+s\}.
\]
If $\gamma+s\ge1$, then $A$ is empty, so assume $\gamma+s<1$.

We first reduce $A$ to a double-sample event.  Draw an independent second sample
$S'=(Z'_1,\ldots,Z'_k)$ and let
\[
    B:=\left\{\exists c\in\cC:\ \widehat q_S(c)\le\gamma
       \text{ and }
       \frac1k\sum_{i=1}^k\mathbf 1[c(Z'_i)=1]>\gamma+s/2\right\}.
\]
Fix $S\in A$ and a witnessing concept $c$.  Conditional on $S$, the number
$Y:=\sum_i\mathbf 1[c(Z'_i)=1]$ is binomial with parameters $k$ and
$p:=p(c)>\gamma+s$.  The lower-tail Chernoff bound gives
\begin{align*}
    \Pr\left[Y\le(\gamma+s/2)k\mid S\right]
    &\le
    \exp\!\left(
        -\frac{k(p-\gamma-s/2)^2}{2p}
    \right)\\
    &\le
    \exp\!\left(-\frac{k s^2}{8(\gamma+s)}\right).
\end{align*}
The second inequality follows because $(p-b)^2/p$ is increasing for $p>b$
and we may take $b=\gamma+s/2$ and $p\ge\gamma+s$.  We will choose $C$ so
that the last display is at most $1/2$.  It then follows by conditioning on
$S$ that
\begin{equation}
    \label{eq:relative-partial-double-sample}
    \Pr_S[A]\le2\Pr_{S,S'}[B].
\end{equation}

It remains to bound $B$.  Condition on the pooled multiset of $2k$ sample
points, retaining the identities of repeated occurrences.  By exchangeability,
the original and second samples form a uniformly random split of these
occurrences into two groups of size $k$.  The quasipolynomial
Sauer--Shelah--Perles lemma for partial concepts gives a family
$\mathcal H$ of total binary labelings of the pooled points such that
\begin{equation}
    \label{eq:relative-partial-disambiguation-size}
    |\mathcal H|\le(2k)^{C_0d\log(2k)}
\end{equation}
for a universal constant $C_0$, and every $c\in\cC$ has an
$h\in\mathcal H$ that agrees with $c$ wherever $c$ is not
$\bot$~\cite[Theorem~4.3]{dhs}.

Suppose a concept $c$ witnesses $B$, and choose such an extension $h$.
On the original sample, every point with $c(Z_i)=0$ must also have
$h(Z_i)=0$.  Therefore the number of original points labeled $1$ by $h$ is
at most the number of nonzero values of $c$, and hence at most $\gamma k$.
On the second sample, every point with $c(Z'_i)=1$ also has $h(Z'_i)=1$, so
more than $(\gamma+s/2)k$ second-sample points are labeled $1$ by $h$.

Fix $h\in\mathcal H$.  Let $J$ be the total number of its $1$-labels among
the pooled occurrences, and let $X$ be the number of these occurrences
assigned to the original sample.  Conditional on the pool, $X$ has the
hypergeometric distribution with population size $2k$, $J$ successes, and
sample size $k$.  The event just described implies
\begin{equation}
    \label{eq:relative-partial-split-event}
    X\le\gamma k,
    \qquad
    J-X>(\gamma+s/2)k.
\end{equation}
Writing $u:=J/(2k)$, we have $\E[X]=uk$.  The first inequality in
\eqref{eq:relative-partial-split-event} gives
$u-X/k\ge u-\gamma$, while the second gives
$u-X/k>\gamma+s/2-u$.  Thus
\begin{equation}
    \label{eq:relative-partial-deviation}
    u-\frac{X}{k}>
    \Delta(u):=\max\{u-\gamma,\,\gamma+s/2-u\}.
\end{equation}
The two arguments of the maximum sum to $s/2$, so
$\Delta(u)\ge s/4$.  We also need a relative, rather than additive, bound.
If $u\ge\gamma+s/4$, then $\Delta(u)=u-\gamma$ and
$(u-\gamma)^2/u$ is increasing in $u$.  If $u\le\gamma+s/4$, then
$\Delta(u)=\gamma+s/2-u$ and $(\gamma+s/2-u)^2/u$ is decreasing in $u$.
Consequently, both cases are minimized at $u=\gamma+s/4$, and
\begin{equation}
    \label{eq:relative-partial-relative-deviation}
    \frac{\Delta(u)^2}{u}
    \ge \frac{(s/4)^2}{\gamma+s/4}
    \ge \frac{s^2}{16(\gamma+s)}.
\end{equation}

If $\Delta(u)>u$, the event in \eqref{eq:relative-partial-deviation} is
impossible because $X\ge0$.  Otherwise, the success indicators in a
hypergeometric sample are negatively associated, so the usual multiplicative
Chernoff lower-tail bound applies and gives
\begin{align}
    \Pr\left[
        X\le\gamma k,\ J-X>(\gamma+s/2)k
        \ \middle|\ \text{pooled sample}
    \right]
    &\le \Pr[X\le(u-\Delta(u))k]\notag\\
    &\le \exp\!\left(-\frac{k\Delta(u)^2}{2u}\right)
     \le \exp\!\left(-\frac{k s^2}{32(\gamma+s)}\right).
    \label{eq:relative-partial-hypergeometric}
\end{align}
This bound is uniform over the pooled sample and over $h\in\mathcal H$.
Taking a union bound using
\eqref{eq:relative-partial-disambiguation-size} therefore yields
\[
    \Pr[B]
    \le
    \exp\!\left(
        C_0d\log^2(2k)-\frac{k s^2}{32(\gamma+s)}
    \right).
\]

For completeness, the choice of $s$ makes the last exponent sufficiently
negative.  If $\gamma\ge s$, then
\[
    \frac{s^2}{\gamma+s}
    \ge\frac{C^2\gamma a}{2\gamma}
    =\frac{C^2a}{2};
\]
if $\gamma<s$, then
\[
    \frac{s^2}{\gamma+s}\ge\frac{s}{2}\ge\frac{Ca}{2}.
\]
Thus $s^2/(\gamma+s)\ge Ca/2$ for $C\ge1$.  Since
$ka=d\log^2(2k)+\log(2/\delta)$, choosing $C$ larger than a suitable
universal constant makes $\Pr[B]\le\delta/2$ and also makes the second-sample
failure probability above at most $1/2$.  Finally,
\eqref{eq:relative-partial-double-sample} gives $\Pr[A]\le\delta$, as
claimed.
\end{proof}

\begin{proof}[Proof of \Cref{thm:relaxed-transfer}]
The assertion is vacuous when $d_{\mathrm{dom}}=\infty$, so assume that
$d_{\mathrm{dom}}<\infty$.
Following the partial-concept reduction of Dwork et
al.~\cite[proof of Theorem~4.1]{dhs}, associate with each $f\in\cF$ the
partial concept
\[
    c_f(D):=
    \begin{cases}
        0, & \err_D(f)\le\tau,\\
        1, & \err_D(f)>\tau+\eta,\\
        \bot, & \tau<\err_D(f)\le\tau+\eta.
    \end{cases}
\]
If this partial class shatters a set, then the same set is
$(\eta/2)$-shattered by $\cF$ at level $\tau+\eta$: a partial label $0$
gives error at most $\tau<\tau+\eta/2$, while a partial label $1$ gives
error greater than $\tau+\eta$.  Hence
\[
    \operatorname{VC}(\{c_f:f\in\cF\})
    \le \Gdim(\cF,\cG,\tau+\eta,\eta/2)
    =d_{\mathrm{dom}}.
\]
The half-margin accommodates the weak inequality
$\err_D(f)\le\tau$ in the relaxed benchmark.

Apply \Cref{lem:relative-partial} on the domain space $\cG$, with $Q=P$
and the partial class just defined.  For every
$f\in\cF_\cD^{(\gamma)}(\tau)$,
\[
    \frac1k\sum_{i=1}^k\mathbf 1[c_f(D_i)\ne0]
    =\frac1k\sum_{i=1}^k
        \mathbf 1[\err_{D_i}(f)>\tau]
    \le\gamma.
\]
Outside the failure event in \Cref{lem:relative-partial}, which has
probability at most $\delta$, no such $f$ can have domain-error rate larger
than the claimed bound.  Indeed,
\[
    \Pr_{D\sim P}[c_f(D)=1]
    =\Pr_{D\sim P}[\err_D(f)>\tau+\eta]
    =\operatorname{Er}_{P,\tau+\eta}(f).
\]
This proves the simultaneous claim, from which the supremum bound follows.
\end{proof}

The supremum in \Cref{thm:relaxed-transfer} is outside the probability over
a new domain.  Thus the theorem controls the bad-domain rate of the worst
fixed member of the relaxed benchmark; it does not claim that
$\Pr_{D\sim P}[\sup_{f\in\cF_\cD^{(\gamma)}(\tau)}
\err_D(f)>\tau+\eta]$ is small.  The stated uniform guarantee is nevertheless
exactly what is needed for a proper, data-dependent learner whose output lies
in the benchmark.

\subsection{Application: sample complexity improvement over Dwork et al.}
\label[subsection]{subsec:meta-sample-complexity}

Suppose $0\le\alpha<\beta<1/2$ and there is a single $f^\star\in\cF$ such
that $\err_D(f^\star)\le\alpha$ for $P$-almost every $D$.  Then an i.i.d.\ tuple
$\cD\sim P^k$ satisfies the $\alpha$-universality condition almost surely.
For any $\gamma\in(0,1/4]$ bounded below by a positive constant,
\Cref{cor:ub-constant-gamma} gives a proper learner using
\[
    N_{\mathrm{lab}} = O\!\left(
        \frac{\beta}{(\beta-\alpha)^2}
        \left[
            d_{\mathrm{VC}}\log\left(
                \frac{e\beta}{(\beta-\alpha)^2}
            \right)
            +\log(1/\delta_{\mathrm{src}})
        \right]
    \right),
    \qquad d_{\mathrm{VC}}:=\operatorname{VC}(\cF),
\]
labeled examples in total across the observed domains and outputting
$\hat f\in\cF_\cD^{(\gamma)}(\beta)$ with probability at least
$1-\delta_{\mathrm{src}}$.

Apply \Cref{thm:relaxed-transfer} with $\tau=\beta$ and domain margin
$\eta$.  Because its event is uniform over the benchmark, it applies to the
data-dependent output $\hat f$.  A union bound shows that, with probability
at least $1-\delta_{\mathrm{src}}-\delta_{\mathrm{dom}}$ over both the
sampled domains and the labeled examples,
\[
    \operatorname{Er}_{P,\beta+\eta}(\hat f)
    \le \gamma+C\left(
        \sqrt{\gamma a_{\mathrm{dom}}}+a_{\mathrm{dom}}
    \right),
\]
where
\[
    a_{\mathrm{dom}}
    :=\frac{
        \Gdim(\cF,\cG,\beta+\eta,\eta/2)\log^2(2k)
        +\log(2/\delta_{\mathrm{dom}})
    }{k}.
\]

In the realizable case $\alpha=0$, this becomes
\[
    N_{\mathrm{lab}} = O\!\left(
        \frac{1}{\beta}
        \left[
            d_{\mathrm{VC}}\log(e/\beta)
            +\log(1/\delta_{\mathrm{src}})
        \right]
    \right).
\]

The number $k$ of observed domains and the number $N_{\mathrm{lab}}$ of
labeled examples play different roles.  The former must still be large enough
to make $a_{\mathrm{dom}}$ small and is therefore governed by the domain
shattering dimension, as in the domain-sample bounds of Dwork et
al.~\cite[Theorems~4.1 and~4.4]{dhs}.  The improvement is in
$N_{\mathrm{lab}}$.  Their min--max ERM analysis assumes uniformly accurate
estimates of every hypothesis's error on every observed
domain~\cite[Section~4.1]{dhs}.  Obtaining
estimation accuracy $u$ from empirical errors uses
\[
    O\!\left(\frac{d_{\mathrm{VC}}+\log(k/\delta)}{u^2}\right)
\]
labels from each of the $k$ domains, and hence
$\widetilde O(k[d_{\mathrm{VC}}+\log(1/\delta)])$ labels in total when the
error margins are constant.  In contrast, for constant
$\gamma$, $\alpha$, and $\beta$, our relaxed-benchmark route uses
$\widetilde O(d_{\mathrm{VC}}+\log(1/\delta_{\mathrm{src}}))$ labels in
total, independently of $k$.

This saving comes with the relaxation parameter $\gamma$: the transferred
bad-domain rate contains an additive $\gamma$.  More generally, to obtain a
target bad-domain rate of order $\zeta$, one may take
$\gamma=\Theta(\zeta)$ and choose $k$ so that
$a_{\mathrm{dom}}=O(\zeta)$.  For fixed classification-error margins,
the random-domain batch branch of \Cref{thm:ub} then uses
$\widetilde O((d_{\mathrm{VC}}+\log(1/\delta_{\mathrm{src}}))/\zeta)$
labels in total, still independently of the possibly much larger $k$.  The
comparison above concerns the labeled cost of the empirical min--max ERM
implementation analyzed by Dwork et al.; it is not a lower bound on the
labeled sample complexity of their learning model.

\section*{AI Disclosure}
OpenAI's ChatGPT and Codex assisted with drafting and editing portions of the
exposition, LaTeX typesetting, and the presentation of proofs throughout the
paper. The authors verified the correctness and originality of all content,
including citations and references, independently checked the mathematical
arguments, and take full responsibility for the submission. The problem
formulation and benchmark were developed by the authors.

\bibliographystyle{plainnat}
\bibliography{ref}

\begin{thebibliography}{54}
\providecommand{\natexlab}[1]{#1}
\providecommand{\url}[1]{\texttt{#1}}
\expandafter\ifx\csname urlstyle\endcsname\relax
  \providecommand{\doi}[1]{doi: #1}\else
  \providecommand{\doi}{doi: \begingroup \urlstyle{rm}\Url}\fi

\bibitem[Aliakbarpour et~al.(2024)Aliakbarpour, Bairaktari, Brown, Smith,
  Srebro, and Ullman]{aliakbarpour2024metalearning}
Maryam Aliakbarpour, Konstantina Bairaktari, Gavin Brown, Adam Smith, Nathan
  Srebro, and Jonathan Ullman.
\newblock Metalearning with very few samples per task.
\newblock In \emph{The Thirty Seventh Annual Conference on Learning Theory},
  pages 46--93. PMLR, 2024.

\bibitem[Alon et~al.(2024)Alon, Hanneke, Moran, and Shalit]{alon2024erm}
Yannay Alon, Steve Hanneke, Shay Moran, and Uri Shalit.
\newblock On the erm principle in meta-learning.
\newblock \emph{arXiv preprint arXiv:2411.17898}, 2024.

\bibitem[Arjovsky et~al.(2019)Arjovsky, Bottou, Gulrajani, and
  Lopez-Paz]{arjovsky2019invariant}
Martin Arjovsky, L{\'e}on Bottou, Ishaan Gulrajani, and David Lopez-Paz.
\newblock Invariant risk minimization.
\newblock \emph{arXiv preprint arXiv:1907.02893}, 2019.

\bibitem[Awasthi et~al.(2023)Awasthi, Haghtalab, and Zhao]{awasthi2023sample}
Pranjal Awasthi, Nika Haghtalab, and Eric Zhao.
\newblock Open problem: The sample complexity of multi-distribution learning
  for {VC} classes.
\newblock In \emph{Proceedings of the Thirty-Sixth Conference on Learning
  Theory}, volume 195 of \emph{Proceedings of Machine Learning Research}, pages
  5943--5949. PMLR, 2023.
\newblock URL \url{https://proceedings.mlr.press/v195/awasthi23a.html}.

\bibitem[Balcan et~al.(2015)Balcan, Blum, and Vempala]{balcan2015efficient}
Maria-Florina Balcan, Avrim Blum, and Santosh Vempala.
\newblock Efficient representations for lifelong learning and autoencoding.
\newblock In \emph{Conference on Learning Theory}, pages 191--210. PMLR, 2015.

\bibitem[Balcan et~al.(2021)Balcan, Khodak, Sharma, and
  Talwalkar]{balcan2021learning}
Maria-Florina~F Balcan, Mikhail Khodak, Dravyansh Sharma, and Ameet Talwalkar.
\newblock Learning-to-learn non-convex piecewise-lipschitz functions.
\newblock \emph{Advances in Neural Information Processing Systems},
  34:\penalty0 15056--15069, 2021.

\bibitem[Ben-David et~al.(2010)Ben-David, Blitzer, Crammer, Kulesza, Pereira,
  and Vaughan]{ben2010theory}
Shai Ben-David, John Blitzer, Koby Crammer, Alex Kulesza, Fernando Pereira, and
  Jennifer~Wortman Vaughan.
\newblock A theory of learning from different domains.
\newblock \emph{Machine learning}, 79\penalty0 (1):\penalty0 151--175, 2010.

\bibitem[Blanchard et~al.(2011)Blanchard, Lee, and
  Scott]{blanchard2011generalizing}
Gilles Blanchard, Gyemin Lee, and Clayton Scott.
\newblock Generalizing from several related classification tasks to a new
  unlabeled sample.
\newblock \emph{Advances in neural information processing systems}, 24, 2011.

\bibitem[Blum et~al.(2017)Blum, Haghtalab, Procaccia, and
  Qiao]{blum2017collaborative}
Avrim Blum, Nika Haghtalab, Ariel Procaccia, and Mingda Qiao.
\newblock Collaborative pac learning.
\newblock \emph{Advances in Neural Information Processing Systems}, 30, 2017.

\bibitem[Chen et~al.(2018)Chen, Zhang, and Zhou]{chen2018tight}
Jiecao Chen, Qin Zhang, and Yuan Zhou.
\newblock Tight bounds for collaborative pac learning via multiplicative
  weights.
\newblock \emph{Advances in neural information processing systems}, 31, 2018.

\bibitem[Collins et~al.(2021)Collins, Hassani, Mokhtari, and
  Shakkottai]{collins2021exploiting}
Liam Collins, Hamed Hassani, Aryan Mokhtari, and Sanjay Shakkottai.
\newblock Exploiting shared representations for personalized federated
  learning.
\newblock In \emph{International conference on machine learning}, pages
  2089--2099. PMLR, 2021.

\bibitem[Cover and Thomas(2006)]{coverThomas2006}
Thomas~M. Cover and Joy~A. Thomas.
\newblock \emph{Elements of Information Theory}.
\newblock Wiley-Interscience, Hoboken, NJ, 2 edition, 2006.
\newblock \doi{10.1002/047174882X}.

\bibitem[Deng et~al.(2020)Deng, Ding, Dwork, Hong, Parmigiani, Patil, and
  Sur]{rvr}
Zhun Deng, Frances Ding, Cynthia Dwork, Rachel Hong, Giovanni Parmigiani,
  Prasad Patil, and Pragya Sur.
\newblock Representation via representations: Domain generalization via
  adversarially learned invariant representations.
\newblock \emph{CoRR}, abs/2006.11478, 2020.
\newblock URL \url{https://arxiv.org/abs/2006.11478}.

\bibitem[Du et~al.(2021)Du, Hu, Kakade, Lee, and Lei]{du2020few}
Simon~S Du, Wei Hu, Sham~M Kakade, Jason~D Lee, and Qi~Lei.
\newblock Few-shot learning via learning the representation, provably.
\newblock In \emph{International Conference on Learning Representations}, 2021.
\newblock URL \url{https://openreview.net/forum?id=pW2Q2xLwIMD}.

\bibitem[Duchi et~al.(2022)Duchi, Feldman, Hu, and Talwar]{duchi2022subspace}
John~C Duchi, Vitaly Feldman, Lunjia Hu, and Kunal Talwar.
\newblock Subspace recovery from heterogeneous data with non-isotropic noise.
\newblock \emph{Advances in Neural Information Processing Systems},
  35:\penalty0 5854--5866, 2022.

\bibitem[Dwork et~al.(2025)Dwork, Hu, and Shao]{dhs}
Cynthia Dwork, Lunjia Hu, and Han Shao.
\newblock How many domains suffice for domain generalization? {A} tight
  characterization via the domain shattering dimension.
\newblock In D.~Belgrave, C.~Zhang, H.~Lin, R.~Pascanu, P.~Koniusz,
  M.~Ghassemi, and N.~Chen, editors, \emph{Advances in Neural Information
  Processing Systems}, volume 38, Main Conference, pages 115008--115034. Curran
  Associates, Inc., 2025.
\newblock \doi{10.52202/085713-3835}.
\newblock URL
  \url{https://proceedings.neurips.cc/paper_files/paper/2025/file/a6e072cfc12794cba1e861f57be8f4de-Paper-Conference.pdf}.

\bibitem[Ganin et~al.(2016)Ganin, Ustinova, Ajakan, Germain, Larochelle,
  Laviolette, Marchand, and Lempitsky]{ganin2016domain}
Yaroslav Ganin, Evgeniya Ustinova, Hana Ajakan, Pascal Germain, Hugo
  Larochelle, Fran{\c{c}}ois Laviolette, Mario Marchand, and Victor Lempitsky.
\newblock Domain-adversarial training of neural networks.
\newblock \emph{Journal of Machine Learning Research}, 17\penalty0
  (59):\penalty0 1--35, 2016.
\newblock URL \url{http://jmlr.org/papers/v17/15-239.html}.

\bibitem[Garg et~al.(2021)Garg, Kalai, Ligett, and Wu]{garg2021learn}
Vikas Garg, Adam~Tauman Kalai, Katrina Ligett, and Steven Wu.
\newblock Learn to expect the unexpected: Probably approximately correct domain
  generalization.
\newblock In \emph{International Conference on Artificial Intelligence and
  Statistics}, pages 3574--3582. PMLR, 2021.

\bibitem[Gong et~al.(2016)Gong, Zhang, Liu, Tao, Glymour, and
  Sch{\"o}lkopf]{gong2016domain}
Mingming Gong, Kun Zhang, Tongliang Liu, Dacheng Tao, Clark Glymour, and
  Bernhard Sch{\"o}lkopf.
\newblock Domain adaptation with conditional transferable components.
\newblock In \emph{International conference on machine learning}, pages
  2839--2848. PMLR, 2016.

\bibitem[Gouk et~al.(2022)Gouk, Bohdal, Li, and
  Hospedales]{gouk2022limitations}
Henry Gouk, Ondrej Bohdal, Da~Li, and Timothy Hospedales.
\newblock On the limitations of general purpose domain generalisation methods.
\newblock \emph{arXiv preprint arXiv:2202.00563}, 2022.

\bibitem[Haghtalab et~al.(2022)Haghtalab, Jordan, and
  Zhao]{haghtalab2022demand}
Nika Haghtalab, Michael Jordan, and Eric Zhao.
\newblock On-demand sampling: Learning optimally from multiple distributions.
\newblock \emph{Advances in Neural Information Processing Systems},
  35:\penalty0 406--419, 2022.

\bibitem[Hanashiro et~al.(2026)Hanashiro, Shetty, and
  Jaillet]{DBLP:conf/colt/HanashiroSJ26}
Rafael Hanashiro, Abhishek Shetty, and Patrick Jaillet.
\newblock Is multi-distribution learning as easy as {PAC} learning: Sharp rates
  with bounded label noise.
\newblock In Steve Hanneke and Tor Lattimore, editors, \emph{The Thirty Ninth
  Annual Conference on Learning Theory, June 29 - July 3, 2026, San Diego,
  California, {USA}}, volume 336 of \emph{Proceedings of Machine Learning
  Research}, pages 3109--3142. {PMLR}, 2026.
\newblock URL \url{https://proceedings.mlr.press/v336/hanashiro26a.html}.

\bibitem[Khodak et~al.(2019)Khodak, Balcan, and Talwalkar]{khodak2019adaptive}
Mikhail Khodak, Maria-Florina~F Balcan, and Ameet~S Talwalkar.
\newblock Adaptive gradient-based meta-learning methods.
\newblock \emph{Advances in neural information processing systems}, 32, 2019.

\bibitem[Khodak et~al.(2023)Khodak, Osadchiy, Harris, Balcan, Levy, Meir, and
  Wu]{khodak2023meta}
Misha Khodak, Ilya Osadchiy, Keegan Harris, Maria-Florina~F Balcan, Kfir~Y
  Levy, Ron Meir, and Steven~Z Wu.
\newblock Meta-learning adversarial bandit algorithms.
\newblock \emph{Advances in Neural Information Processing Systems},
  36:\penalty0 35441--35471, 2023.

\bibitem[Kim et~al.(2022)Kim, Kern, Goldwasser, Kreuter, and
  Reingold]{kim2022universal}
Michael~P Kim, Christoph Kern, Shafi Goldwasser, Frauke Kreuter, and Omer
  Reingold.
\newblock Universal adaptability: Target-independent inference that competes
  with propensity scoring.
\newblock \emph{Proceedings of the National Academy of Sciences}, 119\penalty0
  (4):\penalty0 e2108097119, 2022.

\bibitem[Kong et~al.(2020)Kong, Somani, Song, Kakade, and Oh]{kong2020meta}
Weihao Kong, Raghav Somani, Zhao Song, Sham Kakade, and Sewoong Oh.
\newblock Meta-learning for mixed linear regression.
\newblock In \emph{International Conference on Machine Learning}, pages
  5394--5404. PMLR, 2020.

\bibitem[Larsen et~al.(2024)Larsen, Montasser, and
  Zhivotovskiy]{larsen2024derandomizing}
Kasper~Green Larsen, Omar Montasser, and Nikita Zhivotovskiy.
\newblock Derandomizing multi-distribution learning.
\newblock \emph{Advances in Neural Information Processing Systems},
  37:\penalty0 94246--94264, 2024.

\bibitem[Lattimore and Szepesv{\'a}ri(2020)]{lattimoreSzepesvari2020}
Tor Lattimore and Csaba Szepesv{\'a}ri.
\newblock \emph{Bandit Algorithms}.
\newblock Cambridge University Press, 2020.
\newblock \doi{10.1017/9781108571401}.

\bibitem[Madras et~al.(2018)Madras, Creager, Pitassi, and
  Zemel]{madras2018learning}
David Madras, Elliot Creager, Toniann Pitassi, and Richard Zemel.
\newblock Learning adversarially fair and transferable representations.
\newblock In \emph{International Conference on Machine Learning}, pages
  3384--3393. PMLR, 2018.

\bibitem[Maity et~al.(2024)Maity, Dutta, Terhorst, Sun, and
  Banerjee]{maity2024linear}
Subha Maity, Diptavo Dutta, Jonathan Terhorst, Yuekai Sun, and Moulinath
  Banerjee.
\newblock A linear adjustment-based approach to posterior drift in transfer
  learning.
\newblock \emph{Biometrika}, 111\penalty0 (1):\penalty0 31--50, 2024.

\bibitem[Maurer et~al.(2016)Maurer, Pontil, and
  Romera-Paredes]{maurer2016benefit}
Andreas Maurer, Massimiliano Pontil, and Bernardino Romera-Paredes.
\newblock The benefit of multitask representation learning.
\newblock \emph{Journal of Machine Learning Research}, 17\penalty0
  (81):\penalty0 1--32, 2016.

\bibitem[Mohri et~al.(2019)Mohri, Sivek, and Suresh]{mohri2019agnostic}
Mehryar Mohri, Gary Sivek, and Ananda~Theertha Suresh.
\newblock Agnostic federated learning.
\newblock In \emph{International conference on machine learning}, pages
  4615--4625. PMLR, 2019.

\bibitem[Montasser et~al.(2024)Montasser, Shao, and
  Abbe]{montasser2024transformation}
Omar Montasser, Han Shao, and Emmanuel Abbe.
\newblock Transformation-invariant learning and theoretical guarantees for ood
  generalization.
\newblock \emph{Advances in Neural Information Processing Systems},
  37:\penalty0 108649--108673, 2024.

\bibitem[Muandet et~al.(2013)Muandet, Balduzzi, and
  Sch{\"o}lkopf]{muandet2013domain}
Krikamol Muandet, David Balduzzi, and Bernhard Sch{\"o}lkopf.
\newblock Domain generalization via invariant feature representation.
\newblock In \emph{International conference on machine learning}, pages 10--18.
  PMLR, 2013.

\bibitem[Nguyen and Zakynthinou(2018)]{nguyen2018improved}
Huy Nguyen and Lydia Zakynthinou.
\newblock Improved algorithms for collaborative pac learning.
\newblock \emph{Advances in Neural Information Processing Systems}, 31, 2018.

\bibitem[Patil and Parmigiani(2018)]{patil2018training}
Prasad Patil and Giovanni Parmigiani.
\newblock Training replicable predictors in multiple studies.
\newblock \emph{Proceedings of the National Academy of Sciences}, 115\penalty0
  (11):\penalty0 2578--2583, 2018.

\bibitem[Peng(2024)]{peng2024sample}
Binghui Peng.
\newblock The sample complexity of multi-distribution learning.
\newblock In \emph{The Thirty Seventh Annual Conference on Learning Theory},
  pages 4185--4204. PMLR, 2024.

\bibitem[Rosenfeld et~al.(2021)Rosenfeld, Ravikumar, and
  Risteski]{rosenfeld2020risks}
Elan Rosenfeld, Pradeep Ravikumar, and Andrej Risteski.
\newblock The risks of invariant risk minimization.
\newblock In \emph{International Conference on Learning Representations}, 2021.
\newblock URL \url{https://openreview.net/forum?id=BbNIbVPJ-42}.

\bibitem[Sagawa et~al.(2020)Sagawa, Koh, Hashimoto, and
  Liang]{sagawa2019distributionally}
Shiori Sagawa, Pang~Wei Koh, Tatsunori~B Hashimoto, and Percy Liang.
\newblock Distributionally robust neural networks for group shifts: On the
  importance of regularization for worst-case generalization.
\newblock In \emph{International Conference on Learning Representations}, 2020.
\newblock URL \url{https://openreview.net/forum?id=ryxGuJrFvS}.

\bibitem[Sch{\"o}lkopf et~al.(2012)Sch{\"o}lkopf, Janzing, Peters, Sgouritsa,
  Zhang, and Mooij]{scholkopf2012causal}
Bernhard Sch{\"o}lkopf, Dominik Janzing, Jonas Peters, Eleni Sgouritsa, Kun
  Zhang, and Joris Mooij.
\newblock On causal and anticausal learning.
\newblock \emph{arXiv preprint arXiv:1206.6471}, 2012.

\bibitem[Shao et~al.(2022)Shao, Montasser, and Blum]{shao2022theory}
Han Shao, Omar Montasser, and Avrim Blum.
\newblock A theory of pac learnability under transformation invariances.
\newblock \emph{Advances in Neural Information Processing Systems},
  35:\penalty0 13989--14001, 2022.

\bibitem[Teterwak et~al.(2025)Teterwak, Saito, Tsiligkaridis, Saenko, and
  Plummer]{teterwak2025erm++}
Piotr Teterwak, Kuniaki Saito, Theodoros Tsiligkaridis, Kate Saenko, and Bryan
  Plummer.
\newblock Erm++: An improved baseline for domain generalization.
\newblock In \emph{Proceedings of the Winter Conference on Applications of
  Computer Vision}, pages 8514--8524, 2025.

\bibitem[Thekumparampil et~al.(2021)Thekumparampil, Jain, Netrapalli, and
  Oh]{thekumparampil2021sample}
Kiran~Koshy Thekumparampil, Prateek Jain, Praneeth Netrapalli, and Sewoong Oh.
\newblock Sample efficient linear meta-learning by alternating minimization.
\newblock \emph{arXiv preprint arXiv:2105.08306}, 2021.

\bibitem[Tripuraneni et~al.(2020)Tripuraneni, Jordan, and
  Jin]{tripuraneni2020theory}
Nilesh Tripuraneni, Michael Jordan, and Chi Jin.
\newblock On the theory of transfer learning: The importance of task diversity.
\newblock \emph{Advances in neural information processing systems},
  33:\penalty0 7852--7862, 2020.

\bibitem[Tripuraneni et~al.(2021)Tripuraneni, Jin, and
  Jordan]{tripuraneni2021provable}
Nilesh Tripuraneni, Chi Jin, and Michael Jordan.
\newblock Provable meta-learning of linear representations.
\newblock In \emph{International conference on machine learning}, pages
  10434--10443. Pmlr, 2021.

\bibitem[Vapnik and Chervonenkis(1971)]{vapnikChervonenkis1971}
V.~N. Vapnik and A.~Ya. Chervonenkis.
\newblock On the uniform convergence of relative frequencies of events to their
  probabilities.
\newblock \emph{Theory of Probability and Its Applications}, 16\penalty0
  (2):\penalty0 264--280, 1971.
\newblock \doi{10.1137/1116025}.

\bibitem[Wang et~al.(2022)Wang, Lan, Liu, Ouyang, Qin, Lu, Chen, Zeng, and
  Yu]{wang2022generalizing}
Jindong Wang, Cuiling Lan, Chang Liu, Yidong Ouyang, Tao Qin, Wang Lu, Yiqiang
  Chen, Wenjun Zeng, and Philip~S Yu.
\newblock Generalizing to unseen domains: A survey on domain generalization.
\newblock \emph{IEEE transactions on knowledge and data engineering},
  35\penalty0 (8):\penalty0 8052--8072, 2022.

\bibitem[Wang and Ning(2025)]{wang2025transfer}
Xiaohan Wang and Yang Ning.
\newblock Transfer learning for classification under decision rule drift with
  application to optimal individualized treatment rule estimation.
\newblock \emph{arXiv preprint arXiv:2508.20942}, 2025.

\bibitem[Zhang et~al.(2015)Zhang, Gong, and Sch{\"o}lkopf]{zhang2015multi}
Kun Zhang, Mingming Gong, and Bernhard Sch{\"o}lkopf.
\newblock Multi-source domain adaptation: A causal view.
\newblock In \emph{Proceedings of the AAAI Conference on Artificial
  Intelligence}, volume~29, 2015.

\bibitem[Zhang et~al.(2025)Zhang, Zhan, Chen, Du, and
  Lee]{DBLP:journals/jacm/ZhangZCDL25}
Zihan Zhang, Wenhao Zhan, Yuxin Chen, Simon~S. Du, and Jason~D. Lee.
\newblock Optimal multi-distribution learning.
\newblock \emph{J. {ACM}}, 72\penalty0 (5):\penalty0 36:1--36:71, 2025.
\newblock \doi{10.1145/3760256}.
\newblock URL \url{https://doi.org/10.1145/3760256}.

\bibitem[Zhou et~al.(2022)Zhou, Liu, Qiao, Xiang, and Loy]{zhou2022domain}
Kaiyang Zhou, Ziwei Liu, Yu~Qiao, Tao Xiang, and Chen~Change Loy.
\newblock Domain generalization: A survey.
\newblock \emph{IEEE transactions on pattern analysis and machine
  intelligence}, 45\penalty0 (4):\penalty0 4396--4415, 2022.

\bibitem[Zhu et~al.(2022)Zhu, Lu, Xiao, Han, Luo, and
  Harrison]{zhu2022localized}
Wei Zhu, Le~Lu, Jing Xiao, Mei Han, Jiebo Luo, and Adam~P. Harrison.
\newblock Localized adversarial domain generalization.
\newblock In \emph{2022 IEEE/CVF Conference on Computer Vision and Pattern
  Recognition (CVPR)}, pages 7098--7108, 2022.
\newblock \doi{10.1109/CVPR52688.2022.00697}.

\bibitem[Zhu et~al.(2024)Zhu, Zhang, Gangrade, and Scott]{zhu2024label}
Yilun Zhu, Jianxin Zhang, Aditya Gangrade, and Clayton Scott.
\newblock Label noise: Ignorance is bliss.
\newblock \emph{Advances in neural information processing systems},
  37:\penalty0 116575--116616, 2024.

\bibitem[Zhu et~al.(2025)Zhu, Deng, Shi, Gangrade, and Scott]{zhu2025domain}
Yilun Zhu, Naihao Deng, Naichen Shi, Aditya Gangrade, and Clayton Scott.
\newblock Domain generalization under posterior drift.
\newblock \emph{arXiv preprint arXiv:2510.04441}, 2025.

\end{thebibliography}

\appendix

\section{Useful Results from Information Theory and Learning Theory}
\label[appendix]{sec:tools}

\subsection{Divergences and Information Inequalities}

All logarithms are natural.  We state the results for discrete spaces; the
same arguments extend to general measurable spaces by replacing sums with
the corresponding integrals.  We follow the standard information-theoretic conventions
of \cite[Chapter~2]{coverThomas2006}; for the corresponding sequential and
adaptive formulations, see \cite[Chapter~15]{lattimoreSzepesvari2020}.

\begin{definition}[Relative entropy and binary relative entropy]
For distributions $P,Q$ on $\mathcal Z$, define
\[
    \mathrm{KL}(P\|Q):=\sum_{z\in\mathcal Z}P(z)\log\frac{P(z)}{Q(z)},
\]
with the usual conventions that $P(x)\log\frac{P(x)}{Q(x)}$ is defined to be $0$ whenever $P(x)=0$.
For $p,q\in[0,1]$, write
\[
    \kl(p\|q)
    :=p\log\frac pq+(1-p)\log\frac{1-p}{1-q}.
\]
\end{definition}

Relative entropy is nonnegative and vanishes exactly when the two
distributions agree. We repeatedly use its chain rule: for joint distributions
$P_{XY}$ and $Q_{XY}$,
\begin{equation}
    \label{eq:kl-chain-rule}
    \mathrm{KL}(P_{XY}\|Q_{XY})
    =\mathrm{KL}(P_X\|Q_X)
     +\E_{X\sim P_X}\mathrm{KL}(P_{Y\mid X}\|Q_{Y\mid X}).
\end{equation}
In particular, relative entropy is additive for product distributions. The
same chain rule applies to adaptive transcripts. If an observation is drawn
at round $t$ only on the event that the algorithm has not stopped, its
conditional divergence contributes its one-round divergence multiplied by
the probability of reaching round $t$. This is the stopped chain rule used in
\eqref{eq:hidden-stopped-kl}.

For a random variable $V$ and transcript $T$, mutual information is
\[
    I(V;T):=\mathrm{KL}(P_{VT}\|P_VP_T).
\]
For another random variable $S$, conditional mutual information is
\[
    I(V;T\mid S)
    :=\E_{S}\mathrm{KL}
       (P_{VT\mid S}\|P_{V\mid S}P_{T\mid S}).
\]
Suppose $V\sim\operatorname{Unif}([M])$, and suppose that $T\mid(V=i)\sim P_i$ for each $i\in[M]$. Then the marginal distribution of $T$ is $\overline P=M^{-1}\sum_iP_i$ and
\[
    I(V;T)=\frac1M\sum_{i=1}^M\mathrm{KL}(P_i\|\overline P).
\]
For every reference distribution $Q$,
\begin{equation}
    \label{eq:reference-distribution-decomposition}
    \frac1M\sum_{i=1}^M\mathrm{KL}(P_i\|Q)
    =I(V;T)+\mathrm{KL}(\overline P\|Q),
\end{equation}
and hence
$I(V;T)\le M^{-1}\sum_i\mathrm{KL}(P_i\|Q)$

\begin{definition}[Channel]
A (randomized) channel $K$ from $\mathcal Z$ to $\mathcal W$ assigns to each
$z\in\mathcal Z$ a distribution $K(\,\cdot\mid z)$ on $\mathcal W$. If the
input has distribution $P$, then $KP$ denotes the output distribution,
given in the discrete case by
\[
    (KP)(w):=\sum_{z\in\mathcal Z}P(z)K(w\mid z).
\]
A deterministic map is the special case in which $K(\,\cdot\mid z)$ is a
point mass for every $z$. On general measurable spaces, a channel is also
called a Markov kernel.
\end{definition}

The following standard data-processing inequalities are stated in the form
we use below; see \cite[Section~2.8]{coverThomas2006}.
\begin{lemma}[{Data processing inequalities;
    \citealp[Section~2.8]{coverThomas2006}}]
    \label[lemma]{lem:data-processing}
    Let $K$ be a possibly randomized channel. Then
    \[
        \mathrm{KL}(KP\|KQ)\le\mathrm{KL}(P\|Q).
    \]
    Consequently, if $\phi:\mathcal Z\to\{0,1\}$ and
    $p=P[\phi=1]$, $q=Q[\phi=1]$, then
    \[
        \mathrm{KL}(P\|Q)\ge\kl(p\|q).
    \]
    If $V\to T\to W$ is a Markov chain, then
    \[
        I(V;W)\le I(V;T).
    \]
\end{lemma}

We also use Fano's inequality in its standard finite-hypothesis form
\cite[Section~2.10]{coverThomas2006}.
\begin{lemma}[{Fano's inequality;
    \citealp[Section~2.10]{coverThomas2006}}]
    \label[lemma]{lem:fano}
    Let $V$ be uniform on $[M]$, and let $\widehat V$ be any estimator based
    on an observation $T$. If
    $p_{\mathrm e}:=\Pr[\widehat V\ne V]$, then
    \begin{equation}
        \label{eq:fano}
        I(V;T)\ge
        \log M-h_2(p_{\mathrm e})-p_{\mathrm e}\log(M-1),
    \end{equation}
    where $h_2(p):=-p\log p-(1-p)\log(1-p)$. In particular, if
    $p_{\mathrm e}\le1/4$ and $M\ge2$, then
    $I(V;T)=\Omega(\log M)$.
\end{lemma}

We also use the following standard elementary Bernoulli estimates
\cite[Chapter~2]{coverThomas2006}. For
$0\le p<q\le1/2$,
\begin{equation}
    \label{eq:bernoulli-kl-upper}
    \kl(p\|q)
    \le\frac{(q-p)^2}{q(1-q)}
    \le\frac{2(q-p)^2}{q}.
\end{equation}
We also use the integral identities
\begin{equation}
    \label{eq:binary-kl-integrals}
    \kl(q\|p)=
    \begin{cases}
        \displaystyle\int_p^q\frac{q-t}{t(1-t)}\,dt,&p<q,\\[2mm]
        \displaystyle\int_q^p\frac{t-q}{t(1-t)}\,dt,&q<p.
    \end{cases}
\end{equation}
These identities, or Taylor's theorem, yield the quadratic lower bounds used
in the Chernoff calculations in \Cref{sec:ub,app:epsilon-target-access}.

\subsection{Uniform Convergence Bounds}

The following one-sided threshold formulation of the classical relative VC
inequality is the form used in our upper bounds; see
\cite{vapnikChervonenkis1971}. For a $\{0,1\}$-valued class $\mathcal G$,
let $\Pi_{\mathcal G}$ denote its growth function.

\begin{lemma}[VC threshold separation;
    \citealp{vapnikChervonenkis1971}]
    \label[lemma]{lem:vc-threshold-separation}
    Let $\mathcal G$ be a class of $\{0,1\}$-valued functions with VC
    dimension at most $d$, let $P$ be a distribution, and let $\widehat P_m$
    be the empirical measure of $m$ i.i.d.\ observations from $P$. Fix
    $0\le a<b<1$, set
    \[
        q:=\frac{a+b}{2},
        \qquad
        \Lambda(a,b):=\frac{b(1-a)}{(b-a)^2}.
    \]
    There are universal constants $c,C>0$ such that
    \begin{align}
        \Pr\bigl[\exists g\in\mathcal G:\,
          Pg\le a,\ \widehat P_mg>q\bigr]
        &\le C\Pi_{\mathcal G}(2m)e^{-cm/\Lambda(a,b)},
        \label{eq:vc-threshold-upper}\\
        \Pr\bigl[\exists g\in\mathcal G:\,
          Pg>b,\ \widehat P_mg\le q\bigr]
        &\le C\Pi_{\mathcal G}(2m)e^{-cm/\Lambda(a,b)}.
        \label{eq:vc-threshold-lower}
    \end{align}
    Consequently, for every $\zeta\in(0,1)$, if
    \begin{equation}
        \label{eq:vc-threshold-sample-size}
        m\ge C\Lambda(a,b)
        \left[d\log(e\Lambda(a,b))+\log(1/\zeta)\right],
    \end{equation}
    then, with probability at least $1-\zeta$, simultaneously for every
    $g\in\mathcal G$,
    \begin{equation}
        \label{eq:vc-threshold-separation}
        Pg\le a\implies\widehat P_mg\le q,
        \qquad
        Pg>b\implies\widehat P_mg>q.
    \end{equation}
\end{lemma}

\section{Proofs for the Sample Complexity Upper Bounds}
\label[appendix]{app:upper-bound-proofs}
\label[appendix]{app:subset-certificate}

This appendix proves the two component bounds stated in
\Cref{sec:ub} and then derives the simplifications in
\Cref{cor:ub-uniform,cor:ub-constant-gamma}.  We retain the notation
\(r,B,\kappa,\ell,\rho_{\alpha,\beta}\), and
\(A_{\alpha,\beta}(k,r)\) introduced there.

\subsection{The subset-certificate construction}
\label[subappendix]{app:subset-certificate-analysis}

\begin{proposition}[Subset-certificate construction]
\label[proposition]{prop:ub-subset}
Under the assumptions of \Cref{thm:ub}, there is a nonadaptive proper
learner using
\begin{equation}
\label{eq:upper-bound-subset}
    n
    =
    O\!\left(
        \kappa B
        \left[
            d\log(e\kappa)
            +\log(1/\delta)
            +r\ell
        \right]
    \right)
\end{equation}
labeled source examples.
\end{proposition}

\begin{proof}
    We apply the VC separation lemma, \Cref{lem:vc-threshold-separation}, to
    the class $\cG$ of loss
    functions
    \[
        \mathcal G:=\{(x,y)\mapsto\mathbf 1\{f(x)\ne y\}:f\in\cF\},
    \]
    whose VC dimension equals that of $\cF$. Let
    \[
        \mathcal S_r:=\{S\subseteq[k]:|S|=r\},
        \qquad L:=\log\left(\frac{4|\mathcal S_r|}{\delta}\right).
    \]
   Set $\kappa=\frac{\beta}{(\beta-\alpha)^2}$. Since $1/2<1-\alpha\le 1$, we have
    $\Lambda(\alpha,\beta)=\Theta(\kappa)$. Choose
    \begin{equation}
        \label{eq:upper-pooled-sample-size}
        m=\left\lceil
            C\kappa
            \left[d\log(e\kappa)+L\right]
        \right\rceil
    \end{equation}
    for a sufficiently large universal constant $C$. For each
    $S\in\mathcal S_r$, write
    \[
        D_S:=\frac1r\sum_{i\in S}D_i
    \]
    for the uniform mixture of the source distributions in $S$.

    Each $S\in\mathcal S_r$ is a possible ``elimination certificate'' for
    a candidate hypothesis $\widehat f$: if $\widehat f$ has error greater
    than $\beta$ on every source in $S$, then
    \[
        \err_{D_S}(\widehat f)
        =\frac1r\sum_{i\in S}\err_{D_i}(\widehat f)
        >\beta.
    \]

    The learner uses a fixed number
    \[
        n=\left\lceil C_0\frac{k}{r}m\right\rceil
    \]
    of rounds. On each round it draws $I$ uniformly from $[k]$ and requests
    one labeled example from $D_I$. For $S\in\mathcal S_r$, let $N_S$ be the
    number of rounds for which $I\in S$. Since
    $N_S\sim\operatorname{Binomial}(n,r/k)$, a Chernoff bound and a union bound
    over $\mathcal S_r$ show, for a sufficiently large $C_0$, that
    \begin{equation}
        \label{eq:upper-enough-pooled-samples}
        \Pr[N_S\ge m\text{ for every }S\in\mathcal S_r]
        \ge1-\delta/2.
    \end{equation}
    Indeed, $m\ge C L$, so the failure probability for each fixed $S$ is at
    most $e^{-cm}\le\delta/(2|\mathcal S_r|)$.

    If some $N_S<m$, the learner outputs an arbitrary fixed member of $\cF$.
    Otherwise it proceeds as follows. Conditional on $I\in S$, the labeled
    observation has distribution $D_S$.
    Hence the first $m$ observations whose source indices lie in $S$ are
    i.i.d.\ from $D_S$. Let $\widehat{\err}_S(f)$ be the error of $f$ on these
    $m$ observations. Apply Lemma~\ref{lem:vc-threshold-separation} with
    $a=\alpha$, $b=\beta$, and
    $\zeta=\delta/(2|\mathcal S_r|)$, and take a union bound over
    $S\in\mathcal S_r$. Together with
    \eqref{eq:upper-enough-pooled-samples}, with probability at least
    $1-\delta$ we have, simultaneously for every $S\in\mathcal S_r$ and every
    $f\in\cF$,
    \begin{equation}
        \label{eq:upper-all-subsets-separate}
        \err_{D_S}(f)\le\alpha
        \implies\widehat{\err}_S(f)\le q,
        \qquad
        \err_{D_S}(f)>\beta
        \implies\widehat{\err}_S(f)>q,
    \end{equation}
    where $q=(\alpha+\beta)/2$. Dependence among the samples used for different
    subsets $S$ causes no problem for this union bound.

    By $\alpha$-universality, there is an $f^\star\in\cF$ satisfying
    $\err_{D_i}(f^\star)\le\alpha$ for every $i\in[k]$. Therefore
    $\err_{D_S}(f^\star)\le\alpha$ for every $S\in\mathcal S_r$, and
    \eqref{eq:upper-all-subsets-separate} shows that $f^\star$ belongs to the
    empirical feasible set
    \[
        \widehat{\cF}:=
        \{f\in\cF:\widehat{\err}_S(f)\le q
          \text{ for every }S\in\mathcal S_r\}.
    \]
    The learner outputs any $\widehat f\in\widehat{\cF}$, and outputs the
    arbitrary fixed member of $\cF$ if this set is empty.

    It remains to verify that every empirically feasible hypothesis belongs to
    the relaxed benchmark. Suppose, to the contrary, that
    $\widehat f\notin\cF_\cD^{(\gamma)}(\beta)$. Then $\widehat f$ has error
    greater than $\beta$ on at least
    $r=\lfloor\gamma k\rfloor+1$ sources. Choose a set $S\in\mathcal S_r$ of
    such sources. We have $\err_{D_S}(\widehat f)>\beta$, so
    \eqref{eq:upper-all-subsets-separate} gives
    $\widehat{\err}_S(\widehat f)>q$, contradicting empirical feasibility.
    Thus, on the event \eqref{eq:upper-all-subsets-separate}, the learner
    outputs $\widehat f\in\cF_\cD^{(\gamma)}(\beta)$. This event has
    probability at least $1-\delta$.

    Since the learner is proper, for every target distribution $D'$,
    \[
        \err_{D'}(\widehat f)
        \le\max_{f\in\cF_\cD^{(\gamma)}(\beta)}\err_{D'}(f)
        \le\max_{f\in\cF_\cD^{(\gamma)}(\beta)}\err_{D'}(f)+\varepsilon.
    \]
    This proves both the stated benchmark-membership guarantee and the
    resulting learning guarantee. Finally, $k/r=\Theta(B)$ and
    \[
        \log|\mathcal S_r|=\log\binom{k}{r}=O(r\ell)
    \]
    give \eqref{eq:upper-bound-subset}.
\end{proof}

\subsection{The random-domain batch construction}
\label[subappendix]{app:random-domain-batch-analysis}

\begin{proposition}[Random-domain batch construction]
\label[proposition]{prop:ub-batch}
Under the assumptions of \Cref{thm:ub}, there is a nonadaptive proper
learner using
\begin{equation}
\label{eq:upper-bound-batch}
    n
    =
    O\!\left(
        \kappa B A_{\alpha,\beta}(k,r)
        \left[
            d\log\!\left(
                e\kappa B A_{\alpha,\beta}(k,r)
            \right)
            +\log(1/\delta)
        \right]
    \right)
\end{equation}
labeled source examples.
\end{proposition}

\begin{proof}
    Let
    \[
        g:=\frac{r}{k},
        \qquad
        \ell_g:=\log\frac{e}{g}=\log\frac{ek}{r},
        \qquad
        \tau:=\frac{\alpha+\beta}{2},
    \]
    and set
    \[
        \rho:=
        \begin{cases}
            1+\log(\beta/\alpha),&\alpha>0,\\
            \infty,&\alpha=0,
        \end{cases}
        \qquad
        A_g:=1+\frac{\ell_g}{\rho}.
    \]
    Since $B\le1/g<2B$, we have $\ell_g=\Theta(\ell)$ and
    $A_g=\Theta(A_{\alpha,\beta}(k,r))$.
    Let
    \[
        m_0=\left\lceil
            C\frac{\beta}{(\beta-\alpha)^2}A_g
        \right\rceil
    \]
    for a sufficiently large universal constant $C$. A \emph{domain batch}
    is generated by drawing $I$ uniformly from $[k]$ and then drawing
    $m_0$ independent labeled examples $Z_1,\ldots,Z_{m_0}$ from $D_I$.
    For $f\in\cF$, define the batch-loss function
    \[
        h_f(I,Z_1,\ldots,Z_{m_0})
        :=\mathbf 1\left\{
            \frac1{m_0}\sum_{j=1}^{m_0}
            \mathbf 1\{f(X_j)\ne Y_j\}>\tau
        \right\},
    \]
    and write $R(f)$ for its expectation under the domain-batch
    distribution.
    For a fixed $f\in\cF$ and $i\in[k]$, let
    \[
        \widehat\mu_{m_0}
        :=\frac{1}{m_0}\sum_{j=1}^{m_0}
        \mathbf 1\{f(X_j)\ne Y_j\},
        \qquad
        (X_j,Y_j)\stackrel{\mathrm{i.i.d.}}{\sim}D_i,
    \]
    and write $\mu_i:=\err_{D_i}(f)$. Then
    $\widehat\mu_{m_0}$ is the empirical error of $f$ on the batch, with
    expectation $\mu_i$.

    Write $\Delta:=\beta-\alpha$ and let
    $\kl(q\Vert p)$ denote the binary relative entropy from
    Appendix~\ref{sec:tools}. The Chernoff bounds
    and monotonicity of binomial tails give
    \begin{align*}
        \Pr\left[\widehat\mu_{m_0}>\tau\right]
        &\le \exp\left(-m_0\kl(\tau\Vert\alpha)\right)
        &&\text{when }\mu_i\le\alpha,\\
        \Pr\left[\widehat\mu_{m_0}\le\tau\right]
        &\le \exp\left(-m_0\kl(\tau\Vert\beta)\right)
        &&\text{when }\mu_i>\beta.
    \end{align*}
    The two exponents need not be treated symmetrically. For universal
    constants $c_1,c_2>0$,
    \begin{equation}
        \label{eq:asymmetric-binary-kl}
        \kl(\tau\Vert\alpha)
        \ge c_1\frac{\Delta^2}{\beta}
            \left(1+\log\frac{\beta}{\alpha}\right),
        \qquad
        \kl(\tau\Vert\beta)
        \ge c_2\frac{\Delta^2}{\beta}.
    \end{equation}
    The first inequality is interpreted as an infinite divergence when
    $\alpha=0$. To see these inequalities, use the integral identities
    \eqref{eq:binary-kl-integrals}:
    \[
        \kl(q\Vert p)=
        \begin{cases}
            \displaystyle\int_p^q\frac{q-t}{t(1-t)}\,dt,&p<q,\\[2mm]
            \displaystyle\int_q^p\frac{t-q}{t(1-t)}\,dt,&q<p.
        \end{cases}
    \]
    Since $\tau-\alpha=\beta-\tau=\Delta/2$ and every integration
    variable is at most $\beta$, the corresponding integral gives the lower
    bound $\Delta^2/(8\beta)$ for each divergence. For the first divergence
    it also gives
    \[
        \kl(\tau\Vert\alpha)
        \ge \tau\log\frac{\tau}{\alpha}-(\tau-\alpha).
    \]
    When $\log(\beta/\alpha)$ is at most a universal constant, the quadratic
    lower bound proves the first inequality in
    \eqref{eq:asymmetric-binary-kl}; when it is larger, the last display,
    together with $\tau\ge\beta/2$, supplies the additional logarithmic
    factor. For $\alpha>0$, the identity
    $A_g\rho=\rho+\ell_g\ge\ell_g$ makes the
    first tail probability at most $g/32$ for a sufficiently large $C$; for
    $\alpha=0$, this probability is zero. Since $A_g\ge1$, the second tail
    probability is at most $1/8$. Consequently, for every
    source $i$ and every $f\in\cF$,
    \begin{align}
        \err_{D_i}(f)\le\alpha
        &\implies
        \Pr[h_f=1\mid I=i]\le \frac{g}{32},
        \label{eq:batch-good-domain}\\
        \err_{D_i}(f)>\beta
        &\implies
        \Pr[h_f=1\mid I=i]\ge \frac78.
        \label{eq:batch-bad-domain}
    \end{align}
    In particular, the $\alpha$-universal hypothesis $f^\star$ satisfies
    $R(f^\star)\le g/32$. On the other hand, if
    $f\notin\cF_\cD^{(\gamma)}(\beta)$, then $f$ has error greater than
    $\beta$ on at least $r$ sources, and hence
    \[
        R(f)\ge \frac{7g}{8}>\frac{3g}{4}.
    \]

    Let $\mathcal H:=\{h_f:f\in\cF\}$. On any $T$ domain batches, the
    labeling by $h_f$ is determined by the labeling of the underlying
    $m_0T$ examples by the loss function of $f$. Consequently,
    \begin{equation}
        \label{eq:batch-loss-growth}
        \Pi_{\mathcal H}(2T)
        \le \Pi_{\cG}(2m_0T)
        \le\left(\frac{2em_0T}{d}\right)^d.
    \end{equation}
    Draw
    \[
        T=O\left(
            \frac1g\left[
                d\log\left(\frac{em_0}{g}\right)
                +\log(1/\delta)
            \right]
        \right)
    \]
    independent domain batches, with a sufficiently large constant in the
    big-$O$. Thus, the learner draws independent source indices
    $I_1,\ldots,I_T$ uniformly from $[k]$ and, for each $t$, draws a batch
    $(Z_{t,1},\ldots,Z_{t,m_0})$ from $D_{I_t}$. Define
    \[
        \widehat{\err}_t(f)
        :=\frac1{m_0}\sum_{j=1}^{m_0}
        \mathbf 1\{f(X_{t,j})\ne Y_{t,j}\},
        \qquad
        \widehat R_T(f)
        :=\frac1T\sum_{t=1}^T
        \mathbf 1\{\widehat{\err}_t(f)>\tau\},
    \]
    and let
    \[
        u:=\frac12\left(\frac{g}{32}+\frac{3g}{4}\right)
        =\frac{25g}{64}.
    \]
    The learner outputs
    \begin{equation}
        \label{eq:batch-erm-definition}
        \widehat f\in\argmin_{f\in\cF}\widehat R_T(f),
    \end{equation}
    with ties broken arbitrarily. All source indices and batch sizes are fixed
    before any labeled observations are seen, so the learner is non-adaptive.

    There are two ways in which empirical batch risks could fail to separate
    the universal hypothesis from hypotheses outside the relaxed benchmark:
    \begin{align*}
        E_{\mathrm{FR}}
        &:=\left\{\exists f\in\cF:\
            R(f)\le\frac{g}{32},\ \widehat R_T(f)>u\right\},\\
        E_{\mathrm{FA}}
        &:=\left\{\exists f\in\cF:\
            R(f)>\frac{3g}{4},\ \widehat R_T(f)\le u\right\}.
    \end{align*}
    The event $E_{\mathrm{FR}}$ is a \emph{false rejection}: a genuinely
    low-risk hypothesis, in particular potentially $f^\star$, appears to have
    empirical batch risk above the separating threshold. The event
    $E_{\mathrm{FA}}$ is a \emph{false acceptance}: a high-risk hypothesis,
    which includes every hypothesis outside
    $\cF_\cD^{(\gamma)}(\beta)$, appears to have empirical batch risk below
    the threshold.

    Apply \Cref{lem:vc-threshold-separation}, specifically
    \eqref{eq:vc-threshold-upper} and
    \eqref{eq:vc-threshold-lower}, to $\mathcal H$ with $a=g/32$ and
    $b=3g/4$. Here $\Lambda(a,b)=O(1/g)$, and
    \eqref{eq:batch-loss-growth} gives, for each
    $E\in\{E_{\mathrm{FR}},E_{\mathrm{FA}}\}$,
    \[
        \Pr[E]
        \le C\left(\frac{2em_0T}{d}\right)^d e^{-cgT}
        \le\frac{\delta}{2},
    \]
    where the last inequality follows from the choice of $T$ by a standard
    rearrangement. On the complement of
    $E_{\mathrm{FR}}\cup E_{\mathrm{FA}}$, the universal hypothesis satisfies
    $\widehat R_T(f^\star)\le u$. Hence the minimizing property of $\widehat f$
    gives $\widehat R_T(\widehat f)\le u$. If $\widehat f$ were outside
    $\cF_\cD^{(\gamma)}(\beta)$, then $R(\widehat f)>3g/4$ and the complement
    of $E_{\mathrm{FA}}$ would instead give
    $\widehat R_T(\widehat f)>u$, a contradiction. A union bound therefore
    proves that $\widehat f\in\cF_\cD^{(\gamma)}(\beta)$ with probability at
    least $1-\delta$.

    The learner uses the fixed total budget $n=m_0T$. Substituting $g=r/k$
    and the definition of $m_0$ gives \eqref{eq:upper-bound-batch}. Since the
    output is proper and belongs to the relaxed benchmark, it also satisfies
    the adversarial domain generalization guarantee.
\end{proof}

\subsection{Combining and simplifying the bounds}

\begin{proof}[Proof of \Cref{thm:ub}]
By \Cref{prop:ub-subset}, the subset-certificate learner uses
\[
    O\!\left(
        \kappa B
        [d\log(e\kappa)+\log(1/\delta)+r\ell]
    \right)
\]
samples.  By \Cref{prop:ub-batch}, the random-domain batch learner uses
\[
    O\!\left(
        \kappa B A_{\alpha,\beta}(k,r)
        \left[
            d\log\!\left(e\kappa B A_{\alpha,\beta}(k,r)\right)
            +\log(1/\delta)
        \right]
    \right)
\]
samples.  Both learners are nonadaptive and proper.  Choosing the learner
with the smaller parameter-dependent budget proves
\eqref{eq:upper-bound-main}.
\end{proof}

\begin{proof}[Proof of \Cref{cor:ub-uniform}]
Since \(\rho_{\alpha,\beta}\ge1\) and \(\ell\ge1\),
\[
    A_{\alpha,\beta}(k,r)
    \le 1+\ell\le2\ell.
\]
Substituting this inequality into the random-domain batch branch of
\Cref{thm:ub} proves \eqref{eq:upper-bound-simple}.
\end{proof}

\begin{proof}[Proof of \Cref{cor:ub-constant-gamma}]
If \(\gamma\ge\gamma_0>0\), then
\(r=\lfloor\gamma k\rfloor+1>\gamma k\), and therefore
\[
    B\le\frac{k}{r}<\frac1{\gamma_0},
    \qquad
    \ell\le\log\frac e{\gamma_0},
    \qquad
    A_{\alpha,\beta}(k,r)\le1+\log\frac e{\gamma_0}.
\]
The random-domain batch branch of \Cref{thm:ub} is consequently
\[
    O_{\gamma_0}\!\left(
        \kappa
        \left[
            d\log(e\kappa)+\log(1/\delta)
        \right]
    \right),
\]
which proves \eqref{eq:upper-bound-constant-gamma}.
\end{proof}

\section{Proof of the Hidden Coordinate Lemmas}
\label[appendix]{app:hidden-coordinate-proof}

\begin{proof}[Proof of Lemma~\ref{lem:hidden-coordinate}]
    For each $i\in[M]$, let $\E_i$ denote expectation conditional on
    $I=i$. Since $I$ is uniform on $[M]$,
    \[
        \E[N]=\frac1M\sum_{i=1}^M\E_i[N].
    \]
    If $\E_i[N]=\infty$ for some $i$, then $\E[N]=\infty$ and the claimed
    lower bound is immediate. We may therefore assume without loss of
    generality that $\E_i[N]<\infty$ for every $i\in[M]$.

    Include all of the learner's internal randomness in a random seed $R$, and
    let
    \[
        T=(R,N,v_1,\ldots,v_N),
    \]
    where each $v_i \in \{0,1\}^M$, be the stopped transcript. For $i\in[M]$, write $P_i^T$ for the distribution of $T$
    when the examples have distribution $D_i(\alpha,\beta)$. As a reference
    experiment, let
    \[
        Q=\operatorname{Bernoulli}(\beta)^{\otimes M}
    \]
    be the one-example distribution under which every coordinate has mean
    $\beta$, and let $Q^T$ be the distribution of the stopped transcript obtained by
    running the same learner on i.i.d.\ examples from $Q$.

    The stopped form of the relative-entropy chain rule
    \eqref{eq:kl-chain-rule} gives\footnote{Here, the chain rule is being applied one round at a time.  Letting $T_L$ be the transcript truncated after $L$ possible observations, $\operatorname{KL}(P_i^{T_L}\|Q^{T_L}) =
\sum_{t=1}^L
\Pr[N\ge t|I = i]\,
\mathrm{KL}\bigl(D_i(\alpha,\beta)\,\|\,Q\bigr)$.}
    \begin{align}
        \mathrm{KL}(P_i^T\,\|\,Q^T)
        &= \sum_{t\ge 1}\Pr[N\ge t \,|\,I=i]\,
           \mathrm{KL}\bigl(D_i(\alpha,\beta)\,\|\,Q\bigr) \notag\\
        &= \E_i[N]\,\kl(\alpha\|\beta).                         \label{eq:hidden-stopped-kl}
    \end{align}
    Indeed, conditional on the learner requesting example $t$, that example is
    independent of the past and has distribution $D_i(\alpha,\beta)$ under
    $P_i^T$ and distribution $Q$ under $Q^T$. The learner's stopping
    decisions have the same conditional distribution in the two experiments
    and therefore add no relative
    entropy. Finally, $D_i(\alpha,\beta)$ and $Q$ differ only in coordinate
    $i$, which proves the second equality. The displayed chain rule follows
    directly by truncating at $N\wedge m$ and sending $m\to\infty$.

    Now let $I$ be uniform on $[M]$ and let
    $\overline P^T=M^{-1}\sum_{i=1}^M P_i^T$ be the marginal distribution of $T$.
    The reference-distribution decomposition
    \eqref{eq:reference-distribution-decomposition}, with reference distribution
    $Q^T$, yields
    \begin{align*}
        I(I;T)
        &= \frac1M\sum_{i=1}^M
           \mathrm{KL}(P_i^T\,\|\,\overline P^T)\\
        &\le \frac1M\sum_{i=1}^M
           \mathrm{KL}(P_i^T\,\|\,Q^T)\\
        &= \E[N]\,\kl(\alpha\|\beta),
    \end{align*}
    where the last expectation is over the uniform choice of $I$, the data,
    and the learner's randomness. Since $\widehat I$ is a function of $T$,
    the data processing inequality (Lemma~\ref{lem:data-processing}) and
    Fano's inequality (Lemma~\ref{lem:fano}) give
    \[
        \E[N]\,\kl(\alpha\|\beta)
        \ge I(I;\widehat I)
        \ge \log M-h_2(1/4)-\tfrac14\log(M-1)
        =\Omega(\log M),
    \]
    where $h_2(x):=-x\log x-(1-x)\log(1-x)$. Moreover,
    \eqref{eq:bernoulli-kl-upper} gives
    \[
        \kl(\alpha\|\beta)
        \le \frac{(\beta-\alpha)^2}{\beta(1-\beta)}
        \le \frac{2(\beta-\alpha)^2}{\beta},
    \]
    so
    \begin{equation}
        \label{eq:hidden-size-lower-bound}
        \E[N]\ge\Omega\left(
            \frac{\beta}{(\beta-\alpha)^2}\log M
        \right).
    \end{equation}
\end{proof}

\begin{proof}[Proof of Lemma~\ref{lem:two-coordinate-reference}]
    Include the test's random seed in its stopped transcript $T$. Let
    $R^T,P_1^T,P_2^T$ denote the transcript distributions under $R,P_1,P_2$.
    The stopped KL chain rule gives
    \[
        \mathrm{KL}(R^T\|P_1^T)
        =\E_R[N]\mathrm{KL}(R\|P_1)
        =\E_R[N]\kl(\alpha\|\beta),
    \]
    and, symmetrically,
    \[
        \mathrm{KL}(R^T\|P_2^T)
        =\E_R[N]\kl(\alpha\|\beta).
    \]
    These identities follow because $R$ and $P_1$ differ only in their
    second coordinate, while $R$ and $P_2$ differ only in their first.

    Let $A_i=\{\widehat I=i\}$. Since $A_1$ and $A_2$ are disjoint, choose
    $i\in\{1,2\}$ such that $R^T(A_i)\le1/2$. Then
    $R^T(A_i^c)\ge1/2$, whereas the testing guarantee gives
    $P_i^T(A_i^c)\le\delta$. Binary-event data processing implies
    \begin{align*}
        \E_R[N]\kl(\alpha\|\beta)
        &=\mathrm{KL}(R^T\|P_i^T)\\
        &\ge\kl\bigl(R^T(A_i^c)\|P_i^T(A_i^c)\bigr)\\
        &\ge\kl(1/2\|\delta)
         =\Omega(\log(1/\delta)).
    \end{align*}
    Finally,
    \[
        \kl(\alpha\|\beta)
        \le\frac{(\beta-\alpha)^2}{\beta(1-\beta)}
        \le\frac{2(\beta-\alpha)^2}{\beta},
    \]
    which proves the claim, including when $\alpha=0$.
\end{proof}

\section{Lower Bounds with Target Access}
\label[appendix]{app:epsilon-target-access}

We first prove the noisy-target lower bound, \Cref{thm:m-target-lower},
and then give its realizable counterpart. The latter retains the
zero-benchmark-error property and reaches the $(\log M)/\varepsilon$ scale.

\subsection{Proof of Theorem~\ref{thm:m-target-lower}}

This proof is similar to the lower bound proof when we have no target samples, only this time we choose the target distribution to differ from the reference distribution in a manner that depends on the error parameter $\varepsilon$.  Specifically, in the reference distribution, $X$ is uniform on $\mathcal X$, and $Y$ is an independent fair bit, so $f_t(X)$ is correct with probability $1/2$, where $t$ is the distinguished coordinate. In the target distribution, $X$ is again uniform on $\mathcal X$, but the label is generated from the distinguished coordinate $X_t$: start with $X_t$ and pass it through a binary symmetric channel that leaves it unchanged with probability roughly $1/2 + \varepsilon$, so $f_t(x)$ is correct with probability roughly $1/2+\Omega(\varepsilon)$.
The resulting one-sample KL divergence from the independent-fair-bit reference is $O(\varepsilon^2)$.
\begin{proof}[Proof of Theorem~\ref{thm:m-target-lower}]
    Set $N=kM+1$ and use the coordinate class
    $\cF=\{f_1,\ldots,f_N\}$, where $f_i(x)=x_i$ on
    $\cX=\{0,1\}^N$. Let $\rho\in(\beta,1/2)$ be as in
    \eqref{eq:rho-comparisons}. We use two subfamilies of source-target
    pairs on this same class.

    For the first subfamily, take the balanced-block sources from the proof
    of Theorem~\ref{thm:lb-strict}. Thus a hidden instance is
    $\theta=(t,C_1,\ldots,C_k)$, with $|C_s|=M$, and its benchmark is
    $\{f_t\}$. Put $\sigma_0=\min\{8\varepsilon,1\}$ and define its target by
    drawing $X$ uniformly on $\cX$ and setting
    \[
        Y=X_t\oplus Z,\qquad
        Z\sim\operatorname{Bernoulli}((1-\sigma_0)/2).
    \]
    The benchmark error is $p_\theta=(1-\sigma_0)/2$, and, for every
    classifier $h$,
    \[
        \err_{D_\theta'}(h)=p_\theta
          +\sigma_0\Pr[h(X)\ne f_t(X)].
    \]
    A successful output therefore satisfies
    \[
        \Pr[\widehat f(X)\ne f_t(X)]
        \le \frac{\varepsilon}{\sigma_0}<\frac14;
    \]
    when $\varepsilon\le1/8$ the ratio is $1/8$, and otherwise it equals
    $\varepsilon<1/4$.  Distinct coordinate functions have distance $1/2$
    under the uniform input distribution, so the triangle inequality makes
    $f_t$ the unique coordinate function nearest to $\widehat f$.  Thus every
    successful output identifies $t$ by nearest-coordinate decoding.

    Let $\Pi$ denote the complete transcript, including the target samples,
    all source observations, and the learner's randomness, and let $N_s$
    count the queries to source $s$.
    Put the uniform distribution on the balanced hidden partitions. For a
    fixed block $s$, condition on
    $\mathcal H_s=(\{t\}\cup C_s,(C_{s'})_{s'\ne s})$. Conditional on
    $\mathcal H_s$, the index $t$ is uniform on
    $U_s:=\{t\}\cup C_s$, which has $M+1$ elements. For $u\in U_s$, let
    $P_u^\Pi$ be the conditional distribution of the complete transcript when
    $t=u$.  We compare these distributions with a common rencrence
    distribution $Q_s^\Pi$, obtained by
    running the same learner in the following reference experiment:
    \begin{itemize}
        \item each target input is uniform on $\cX$, but its label is an
        independent fair bit;
        \item on source $s$, all error indicators indexed by $U_s$ have mean
        $\rho$, so there is no exceptional coordinate in $U_s$;
        \item all remaining source observations have their original
        $u$-independent conditional distributions.
    \end{itemize}
    This reference experiment may depend on $\mathcal H_s$, but it does not
    depend on the unknown index $u$.

    For one target observation, the input distribution is the same under
    $P_u^\Pi$ and $Q_s^\Pi$. Conditional on the input, its label is the output
    of a binary symmetric channel with crossover probability
    $(1-\sigma_0)/2$ under $P_u^\Pi$ and is fair under $Q_s^\Pi$. Its KL
    contribution is therefore
    \[
        \kl\left(\frac{1-\sigma_0}{2}\middle\|\frac12\right)
        =\log2-h_2((1-\sigma_0)/2)
        \le C\varepsilon^2.
    \]
    A source query outside block $s$ has identical conditional distribution in the
    two experiments and contributes zero KL\@. At a query to source $s$, the
    distributions differ only at coordinate $u$, whose error mean is $\alpha$ under
    $P_u^\Pi$ and $\rho$ under $Q_s^\Pi$; hence its contribution is
    $\kl(\alpha\|\rho)$. The learner's adaptive source choice is determined
    by the preceding transcript and contributes no KL of its own. Thus the
    adaptive KL chain rule, exactly as in the proof of the Hidden Coordinate Lemma (Lemma~\ref{lem:hidden-coordinate}) Equation~\eqref{eq:hidden-stopped-kl}, and again letting
    $E_i$ denote expectation conditional on $I=i$, for each $i\in[M]$,
    gives
    \[
        \mathrm{KL}(P_u^\Pi\|Q_s^\Pi)
        \le Cm\varepsilon^2
          +\E_u[N_s\mid\mathcal H_s]\kl(\alpha\|\rho).
    \]

    Let $\overline P^\Pi=(M+1)^{-1}\sum_{u\in U_s}P_u^\Pi$ be the conditional
    mixture transcript distribution. Applying the reference-distribution
    decomposition \eqref{eq:reference-distribution-decomposition}, as in the
    proof of Lemma~\ref{lem:hidden-coordinate}, now yields
    \begin{align*}
        I(t;\Pi\mid\mathcal H_s)
        &=\frac1{M+1}\sum_{u\in U_s}
          \mathrm{KL}(P_u^\Pi\|\overline P^\Pi)\\
        &\le\frac1{M+1}\sum_{u\in U_s}
          \mathrm{KL}(P_u^\Pi\|Q_s^\Pi)\\
        &\le Cm\varepsilon^2
          +\E_u[N_s\mid\mathcal H_s]\kl(\alpha\|\rho).
    \end{align*}
    Here the last expectation averages over the uniform choice of
    $u\in U_s$ as well as the learner and the observations.
    Fano's inequality lower-bounds the left side by $c\log M$. Averaging,
    summing over $s$, using $\sum_sN_s=n$, and applying
    $\kl(\alpha\|\rho)\le2(\rho-\alpha)^2/\rho$ gives
    \begin{equation}
        \max\left\{
          \frac{(\beta-\alpha)^2}{\beta}n,
          \;k\varepsilon^2m
        \right\}
        \ge c_1k\log M.
        \label{eq:noisy-target-strengthened-size}
    \end{equation}

    For the confidence term, use the following hidden-block-size
    construction.
    Fix $u\ne v$ and let
    $g:[N]\setminus\{u,v\}\to[k]$ be an owner map with one fiber (pre-image) of size $M-1$ and the other $k-1$ fibers of size $M$.  For each $s\in [k]$, form $\cD^{u,s}$ by making $u$
    exceptional and assigning $v$ to source $s$, and form $\cD^{v,s}$ with
    the roles reversed. Their benchmarks are $\{f_u\}$ and $\{f_v\}$.

    Fix $x^{u,v}$ with $x^{u,v}_u=0$ and $x^{u,v}_v=1$. For
    $w\in\{u,v\}$, associate $\cD^{w,s}$ with the target supported on
    $X=x^{u,v}$ and
    \[
        Y=f_w(X)\oplus Z,\qquad
        Z\sim\operatorname{Bernoulli}(p_1),\qquad
        p_1:=\frac{1-2\varepsilon}{2}.
    \]
    The benchmark error is $p_1$, whereas predicting the wrong label at
    $x^{u,v}$ has error $p_1+2\varepsilon$. Success therefore determines
    whether $w=u$ or $w=v$.

    Define a common reference experiment as follows. Under reference source
    $R_j$, the coordinates in $g^{-1}(j)$ have error mean $\rho$ and all
    others, including $u,v$, have error mean $\alpha$. The reference target
    is supported on $x^{u,v}$ and has an independent fair label. Write
    $Q^\Pi$ for the complete reference transcript distribution, and define
    \[
        A_w:=\{\widehat f(x^{u,v})=f_w(x^{u,v})\},
        \qquad w\in\{u,v\}.
    \]
    Since $f_u(x^{u,v})\ne f_v(x^{u,v})$, the events $A_u$ and $A_v$ are
    disjoint.  Choose $w$ such that $Q^\Pi(A_w)\le1/2$; this same choice
    is used for every source $s$.

    If $P_{w,s}^\Pi$ is the transcript distribution for the valid pair
    $(\cD^{w,s},D_{w,s}')$, the adaptive KL chain rule gives
    \[
        \mathrm{KL}(Q^\Pi\|P_{w,s}^\Pi)
        =m\kl\left(\frac12\middle\|\frac{1-2\varepsilon}{2}\right)
          +\E_{Q^\Pi}[N_s]\kl(\alpha\|\rho)
        \le C'm\varepsilon^2
          +\E_{Q^\Pi}[N_s]\kl(\alpha\|\rho).
    \]
    On the other hand, $Q^\Pi(A_w^c)\ge1/2$ and the success guarantee
    gives $P_{w,s}^\Pi(A_w^c)\le\delta$. Binary-event data processing
    therefore makes the last display at least
    $\kl(1/2\|\delta)=\Omega(\log(1/\delta))$. Summing this inequality over
    $s$, using $\sum_s\E_{Q^\Pi}N_s=n$, and again applying
    \eqref{eq:rho-comparisons} and the Bernoulli KL upper bound yields
    \begin{equation}
        \max\left\{
          \frac{(\beta-\alpha)^2}{\beta}n,
          \;k\varepsilon^2m
        \right\}
        \ge c_2k\log(1/\delta).
        \label{eq:noisy-target-strengthened-confidence}
    \end{equation}
    This includes $\alpha=0$, since $\kl(0\|\rho)<\infty$. Combining the
    last two displays proves \eqref{eq:target-access-epsilon-squared-lower}.
\end{proof}
\subsection{Realizable Target-Access Lower Bound}
Recall the intuition for dependence on $\varepsilon$ in the realizable case, where we require the benchmark error on the target to be zero:  Any candidate function with target error at least $\varepsilon$ is contradicted by a fresh labeled target example with probability at least $\varepsilon$.

\begin{theorem}[Realizable target at the $1/\varepsilon$ scale]
    \label[theorem]{thm:target-access-epsilon}
    Fix $k\ge1$, $0\le\alpha<\beta<1/2$,
    $0<\varepsilon<1/4$, $\delta\in(0,1/4)$, and integers $m\ge0$ and
    $M\ge2$. There is a universal constant $c_0>0$, a hypothesis class
    $\cF$ of size $kM+1$, and a family of $\alpha$-universal source tuples
    $\cD^\theta=(D_1^\theta,\ldots,D_k^\theta)$ with associated target
    distributions $D_\theta'$ such that
    \[
        \max_{f\in\cF_{\cD^\theta}(\beta)}
        \err_{D_\theta'}(f)=0.
    \]
    Suppose a learner receives $m$ i.i.d.\ labeled examples from $D_\theta'$, then
    uses a fixed total budget of $n$ adaptively selected source samples, and
    satisfies, for every hidden instance $\theta$,
    \[
        \Pr\bigl[\err_{D_\theta'}(\widehat f)\le\varepsilon\bigr]
        \ge1-\delta.
    \]
    Then
    \begin{equation}
        \label{eq:target-access-epsilon-lower}
        \max\left\{
            \frac{(\beta-\alpha)^2}{\beta}\,n,
            \;k\varepsilon m
        \right\}
        \ge c_0k\bigl[\log M+\log(1/\delta)\bigr].
    \end{equation}
\end{theorem}

\begin{proof}
    Use the hypothesis class and source tuples from the proof of the Strict Benchmark lower bound (Theorem~\ref{thm:lb-strict}), including the choice of
    $\rho\in(\beta,1/2)$. Thus
    $\cF_{\cD^\theta}(\beta)=\{f_t\}$. Set
    \[
        q:=\min\{8\varepsilon,1\}.
    \]
    Define $D_t'$ by drawing $(X,Y)$ as follows. With probability $1-q$, set
    $(X,Y)=(0^N,0)$, where $N=kM+1$. With probability $q$, draw $X$ uniformly
    from $\{0,1\}^N$ and set $Y=X_t$. Since all coordinate hypotheses vanish
    at $0^N$, by construction we always have $Y=f_t(X)$, and hence
    $\err_{D_t'}(f_t)=0$.

    Let $P_X$ be the input marginal of $D_t'$ and write
    $d_q(g,h):=\Pr_{X\sim P_X}[g(X)\ne h(X)]$. For $i\ne j$,
    \[
        d_q(f_i,f_j)=\frac q2>2\varepsilon.
    \]
    Indeed, this is $4\varepsilon$ when $\varepsilon\le1/8$, and it is
    $1/2>2\varepsilon$ otherwise. If
    $\err_{D_t'}(\widehat f)\le\varepsilon$, then
    $d_q(\widehat f,f_t)\le\varepsilon$, whereas the triangle inequality
    gives
    \[
        d_q(\widehat f,f_j)
        \ge d_q(f_t,f_j)-d_q(\widehat f,f_t)
        >\varepsilon
        \qquad(j\ne t).
    \]
    Thus nearest-neighbor decoding identifies $t$ whenever the learner
    succeeds.

    Put the uniform distribution on the hidden ordered partitions
    $\theta=(t,C_1,\ldots,C_k)$ used in the proof of
    Theorem~\ref{thm:lb-strict}. Fix $s\in[k]$, condition on
    $U_s:=\{t\}\cup C_s$, and condition on all blocks $C_{s'}$ with
    $s'\ne s$. Under this conditioning, $t$ is uniform on the $M+1$ elements
    of $U_s$. Let $T_{\rm tar}$ be the $m$-example target sample and let
    $\Pi$ be the complete transcript, including $T_{\rm tar}$, all source
    observations, and the learner's randomness. For brevity, denote the
    information conditioned on above by
    $\mathcal H_s:=(U_s,(C_{s'})_{s'\ne s})$.

    For $u\in U_s$, let $P_u^\Pi$ be the conditional distribution of the complete
    transcript when $t=u$. Define a common reference distribution
    $Q_s^\Pi$ by running
    the same learner in the following experiment. For each target example,
    first draw the same component indicator
    $B\sim\operatorname{Bernoulli}(q)$ as in the hard instance. If $B=0$,
    return $(0^N,0)$; if $B=1$, draw $X$ uniformly from $\{0,1\}^N$ and draw
    $Y$ as an independent fair bit. On source $s$, give every error indicator
    indexed by $U_s$ mean $\rho$; all remaining source observations retain
    their original $u$-independent conditional distributions. This reference depends
    on $\mathcal H_s$ but not on $u$.

    It is convenient to augment both target experiments by the latent
    component indicator $B$. Conditional on $B=0$, their target observations
    agree. Conditional on $B=1$, their inputs agree and the hard-instance
    label is deterministic given $(X,u)$ whereas the reference label is
    fair. By data processing after discarding $B$, the KL contribution of
    one observed target example is therefore at most $q\log2$.

    Let $N_s$ denote the number of source queries to block $s$.
    A source query outside block $s$ has the same conditional distribution in the two
    experiments and contributes zero KL\@. At a query to source $s$, only the
    error indicator at coordinate $u$ changes, from mean $\alpha$ under
    $P_u^\Pi$ to mean $\rho$ under $Q_s^\Pi$, and hence contributes
    $\kl(\alpha\|\rho)$. The learner's adaptive source choice is determined
    by the preceding transcript and contributes no KL\@. Consequently, the
    adaptive KL chain rule gives
    \[
        \mathrm{KL}(P_u^\Pi\|Q_s^\Pi)
        \le mq\log2
          +\E_u[N_s\mid\mathcal H_s]\kl(\alpha\|\rho).
    \]

    Let $\overline P^\Pi=(M+1)^{-1}\sum_{u\in U_s}P_u^\Pi$. The
    reference-distribution decomposition
    \eqref{eq:reference-distribution-decomposition} yields
    \begin{align*}
        I(t;\Pi\mid\mathcal H_s)
        &=\frac1{M+1}\sum_{u\in U_s}
          \mathrm{KL}(P_u^\Pi\|\overline P^\Pi)\\
        &\le\frac1{M+1}\sum_{u\in U_s}
          \mathrm{KL}(P_u^\Pi\|Q_s^\Pi)\\
        &\le mq\log2
          +\E[N_s\mid\mathcal H_s]\kl(\alpha\|\rho).
    \end{align*}
    Since $q\le8\varepsilon$, this proves directly that
    \begin{equation}
        \label{eq:target-source-information}
        I(t;\Pi\mid U_s,(C_{s'})_{s'\ne s})
        \le 8m\varepsilon\log2
          +\E[N_s\mid U_s,(C_{s'})_{s'\ne s}]
             \kl(\alpha\|\rho).
    \end{equation}
    Since the decoded index has error at most $\delta<1/4$, Fano's inequality
    (Lemma~\ref{lem:fano}) lower-bounds the left-hand side by $c_1\log M$
    for a universal $c_1>0$. Moreover, the Bernoulli estimate
    \eqref{eq:bernoulli-kl-upper} gives
    \[
        \kl(\alpha\|\rho)
        \le \frac{2(\rho-\alpha)^2}{\rho},
    \]
    and \eqref{eq:rho-comparisons} implies
    $\rho/(\rho-\alpha)^2
      =\Omega(\beta/(\beta-\alpha)^2)$. Averaging the conditioning, summing
    over $s\in[k]$, and using $\sum_sN_s=n$ proves the first term in
    \eqref{eq:target-access-epsilon-lower}.

    It remains to prove the strengthened confidence term. We use a second
    subfamily on the same coordinate class. Fix $u\ne v$ and an owner map
    $g:[N]\setminus\{u,v\}\to[k]$ with one fiber of size $M-1$ and all
    other fibers of size $M$. For each $s\in[k]$, construct
    $\cD^{u,s}$ and $\cD^{v,s}$ by making coordinate $u$ (respectively
    $v$) exceptional and assigning the other coordinate to source $s$;
    coordinates governed by $g$ retain their owners. Their benchmark sets
    are $\{f_u\}$ and $\{f_v\}$.

    Choose $x^{u,v}$ with $x^{u,v}_u=0$ and $x^{u,v}_v=1$, and put
    $q_1:=2\varepsilon$. The realizable target for $\cD^{w,s}$,
    $w\in\{u,v\}$, equals $(0^N,0)$ with probability $1-q_1$ and equals
    $(x^{u,v},f_w(x^{u,v}))$ with probability $q_1$. Thus $f_w$ has zero
    target error, while predicting the wrong label at $x^{u,v}$ incurs error
    $q_1=2\varepsilon$. Every successful output therefore identifies $w$.

    Define a common reference source tuple $\mathcal R$ so that under
    reference source $R_j$, the coordinates in $g^{-1}(j)$ have error mean
    $\rho$, and all
    others, including $u,v$, have error mean $\alpha$. Let the reference
    target put all its mass on $(0^N,0)$, and let $Q^\Pi$ be the
    distribution of the complete reference transcript. Define
    \[
        A_w:=\{\widehat f(x^{u,v})=f_w(x^{u,v})\},
        \qquad w\in\{u,v\}.
    \]
    These two events are disjoint.  Choose $w$ such that
    $Q^\Pi(A_w)\le1/2$; this choice is common to every $s$.

    Let $P_{w,s}^\Pi$ be the transcript distribution for the valid
    source--target pair associated with $\cD^{w,s}$. The reference target is absolutely
    continuous with respect to the realizable target, with one-sample
    divergence $\log(1/(1-q_1))=O(\varepsilon)$. Moreover, only queries to
    source $s$ distinguish the source tuples. Hence the adaptive KL chain
    rule gives
    \begin{equation}
        \mathrm{KL}(Q^\Pi\|P_{w,s}^\Pi)
        =m\log\frac1{1-q_1}
          +\E_{Q^\Pi}[N_s]\kl(\alpha\|\rho)
        \le Cm\varepsilon
          +\E_{Q^\Pi}[N_s]\kl(\alpha\|\rho).
        \label{eq:realizable-target-reference-kl}
    \end{equation}
    Since $Q^\Pi(A_w^c)\ge1/2$ and success implies
    $P_{w,s}^\Pi(A_w^c)\le\delta$, binary-event data processing makes
    \eqref{eq:realizable-target-reference-kl} at least
    $\kl(1/2\|\delta)=\Omega(\log(1/\delta))$. Summing over $s$, using
    $\sum_s\E_{Q^\Pi}N_s=n$, and applying
    $\kl(\alpha\|\rho)\le2(\rho-\alpha)^2/\rho$ gives
    \[
        \max\left\{
          \frac{(\beta-\alpha)^2}{\beta}n,
          \;k\varepsilon m
        \right\}
        \ge c_2k\log(1/\delta).
    \]
    This remains valid at $\alpha=0$ because
    $\kl(0\|\rho)<\infty$. Combining it with the previously established
    $k\log M$ bound proves \eqref{eq:target-access-epsilon-lower}.
\end{proof}

The noisy, and generally nonzero, benchmark error in
Theorem~\ref{thm:m-target-lower} is what permits the
$1/\varepsilon^2$ target-information scale. At zero benchmark error, the
following elementary upper bound gives a $1/\varepsilon$ ceiling instead.

\begin{proposition}[Target-only upper bound at zero benchmark error]
    \label[proposition]{prop:zero-target-upper}
    Let $\cF$ be finite, let $\cF_0\subseteq\cF$ be nonempty, and suppose a
    target distribution $D'$ satisfies
    \[
        \max_{f\in\cF_0}\err_{D'}(f)=0.
    \]
    Given
    \[
        m\ge\frac{\log|\cF|+\log(1/\delta)}{\varepsilon}
    \]
    labeled target examples, a learner using no source samples can output
    $\widehat f\in\cF$ such that
    \[
        \Pr[\err_{D'}(\widehat f)\le\varepsilon]\ge1-\delta.
    \]
\end{proposition}

\begin{proof}
    Output any hypothesis in $\cF$ consistent with the target sample. Such a
    hypothesis exists because every member of $\cF_0$ has zero target error.
    For any $f\in\cF$ with $\err_{D'}(f)>\varepsilon$, the probability that
    $f$ is consistent with all $m$ examples is at most
    $(1-\varepsilon)^m\le e^{-m\varepsilon}$. A union bound over $\cF$ proves
    the claim.
\end{proof}

In particular, when $\cF_0$ is the source benchmark and its maximum target
error is zero, the proposition achieves the target-access guarantee without
querying any source. For the classes above, $|\cF|=kM+1$, so a source lower
bound cannot in general persist beyond order
$(\log(kM)+\log(1/\delta))/\varepsilon$ target samples while retaining the
zero-benchmark-error property. In particular, when $k$ is fixed or polynomial
in $M$, this rules out persistence to order $(\log M)/\varepsilon^2$ as
$\varepsilon\to0$.

\subsection{Relaxed-Benchmark Target-Access Lower Bound}
Both lower bounds extend to the relaxed benchmark by exactly the
domain-repetition reduction used in the proof of Theorem~\ref{thm:lb}.

\begin{corollary}[Relaxed target-access lower bounds]
    \label[corollary]{cor:relaxed-target-access}
    Fix $k\ge1$, $\gamma\in[0,1)$, and define
    \[
        r:=\lfloor\gamma k\rfloor+1,
        \qquad
        B:=\lfloor k/r\rfloor.
    \]
    In the setting of Theorem~\ref{thm:target-access-epsilon}, there is a
    class of size $BM+1$ for which every learner satisfying the target-access
    guarantee with the relaxed benchmark $\cF_\cD^{(\gamma)}(\beta)$ obeys
    \[
        \max\left\{
            \frac{(\beta-\alpha)^2}{\beta}\,n,
            \;B\varepsilon m
        \right\}
        \ge c_0B\bigl[\log M+\log(1/\delta)\bigr]
    \]
    for the zero-benchmark targets of
    Theorem~\ref{thm:target-access-epsilon}. For the noisy targets of
    Theorem~\ref{thm:m-target-lower}, the corresponding bound
    is
    \[
        \max\left\{
            \frac{(\beta-\alpha)^2}{\beta}\,n,
            \;B\varepsilon^2m
        \right\}
        \ge c_0B\bigl[\log M+\log(1/\delta)\bigr].
    \]
\end{corollary}

\begin{proof}
    Start from either strict $B$-source construction above. Partition the
    $k$ source indices into $B$ groups of size at least
    $r=\lfloor\gamma k\rfloor+1$ and repeat each strict source throughout its
    group, exactly as in the proof of Theorem~\ref{thm:lb}. If
    $\widetilde\cD$ is the original $B$-tuple and $\cD$ is the repeated
    $k$-tuple, then
    \[
        \cF_\cD^{(\gamma)}(\beta)
        =\cF_{\widetilde\cD}(\beta)=\{f_t\}.
    \]
    The target distribution and its $m$ labeled examples are unchanged. A
    query to any repeated source can be simulated by a fresh query to the
    corresponding original source, including under adaptive source
    selection. The two strict lower bounds with $B$ in place of $k$ therefore
    give the claimed inequalities.
\end{proof}

\begin{proof}[Proof of \Cref{cor:target-access-vc}]
    Apply the noisy-target part of \Cref{cor:relaxed-target-access} with
    \[
        M:=\left\lfloor\frac{2^d}{B}\right\rfloor.
    \]
    The calculation in the proof of \Cref{cor:lb-vc} shows that $M\ge2$,
    that the resulting class of size $BM+1$ has VC dimension at most $d$,
    and that $\log M=\Omega(d)$.  Substituting this last bound into the noisy
    inequality of \Cref{cor:relaxed-target-access} proves the claim.
\end{proof}

\section{Multi-Distribution Learning under Majority Closure}
\label[appendix]{app:mdl-closure}

This appendix isolates the positive results that use a structural closure
assumption. Results outside this appendix do not require it. We use MDL as
shorthand for the multi-distribution learning routine invoked below.

\begin{definition}[Majority closure]
    We call $\cF$ \emph{majority-closed} if, for every positive integer $T$
    and every multiset $f_1,\ldots,f_T\in\cF$, the thresholded pointwise vote
    \[
        x\longmapsto
        \mathbf 1\!\left[\frac1T\sum_{t=1}^T f_t(x)\ge\frac12\right]
    \]
    also belongs to $\cF$.
\end{definition}

This is a strong condition: the required value of $T$ may depend on the
accuracy and confidence parameters, yet the vote must remain in the original
binary hypothesis class.

\subsection{Strict benchmark}

\begin{theorem}[Majority-closure upper bound]
    \label[theorem]{thm:mdl-upper}
    Let $d = \mathrm{VC}(\cF)$. Suppose $\cF$ is majority-closed and the source
    distributions satisfy $\alpha$-universality. If $\beta > 2\alpha$, then
    there is a proper learner that uses
    \[
        \widetilde O\!\left(
            \frac{d + k + \log(1/\delta)}{(\beta-2\alpha)^2}
        \right)
    \]
    source samples and outputs $\hat{f} \in \cF$ satisfying, with probability
    at least $1-\delta$,
    \[
        \max_{i\in[k]} \err_{D_i}(\hat{f}) \le \beta.
    \]
    In particular $\hat{f} \in \cF_\cD(\beta)$, so $\hat{f}$ automatically
    satisfies the guarantee in \Cref{def:main}.
\end{theorem}

\begin{proof}
    Set $\eta := \beta - 2\alpha > 0$.
    The randomized multi-distribution learning algorithm of
    \citet{DBLP:journals/jacm/ZhangZCDL25} gives a finite
    multiset $f_1,\ldots,f_T \in \cF$ satisfying, with
    probability at least $1-\delta$,
    \[
        \max_{i\in[k]}\frac{1}{T}\sum_{t=1}^T \err_{D_i}(f_t)
        \le \alpha + \tfrac{\eta}{2},
    \]
    using $\widetilde O((d+k+\log(1/\delta))/\eta^2)$ source samples.
    Define the deterministic majority vote
    \[
        \hat{f}(x) := \mathbf{1}\!\left[\frac{1}{T}\sum_{t=1}^T f_t(x)
        \ge \tfrac{1}{2}\right].
    \]
    Since $\cF$ is majority-closed, $\hat{f} \in \cF$. For any labeled example
    $(x,y)$, if $\hat{f}(x) \neq y$ then at least half the multiset predicts
    incorrectly, giving
    \[
        \mathbf{1}[\hat{f}(x)\neq y]
        \le \frac{2}{T}\sum_{t=1}^T \mathbf{1}[f_t(x)\neq y].
    \]
    Taking expectation over $(x,y) \sim D_i$:
    \[
        \err_{D_i}(\hat{f})
        \le \frac{2}{T}\sum_{t=1}^T \err_{D_i}(f_t)
        \le 2\alpha + \eta
        = \beta \quad \text{for every } i\in[k].
    \]
    Hence $\hat{f} \in \cF_\cD(\beta)$, so the observation following
    \Cref{def:main} shows that $\hat{f}$ satisfies
    \eqref{eq:error-goal}.
\end{proof}

\subsection{Relaxed benchmark}

The same output also handles every relaxed benchmark, because it violates the
threshold $\beta$ on no source at all.

\begin{theorem}[Majority closure for relaxed benchmarks]
    \label[theorem]{thm:relaxed-mdl}
    Assume $\cF$ is majority-closed and let $d = \mathrm{VC}(\cF)$.
    If $\beta > 2\alpha$, then for every $\gamma \in [0,1]$ there is a proper
    learner that uses
    \[
        \widetilde O\!\left(
            \frac{d + k + \log(1/\delta)}{(\beta-2\alpha)^2}
        \right)
    \]
    source samples and outputs $\hat{f} \in \cF_\cD^{(\gamma)}(\beta)$ with
    probability at least $1-\delta$. Consequently, the same learner satisfies
    the relaxed guarantee in \Cref{def:main-relaxed}.
\end{theorem}

\begin{proof}
    By \Cref{thm:mdl-upper}, the MDL procedure returns a proper hypothesis
    $\hat{f} \in \cF$ such that, with probability at least $1-\delta$,
    \[
        \max_{i \in [k]} \err_{D_i}(\hat{f}) \le \beta.
    \]
    Hence the set of sources on which $\hat{f}$ violates the threshold
    $\beta$ is empty, and therefore
    $\hat{f} \in \cF_\cD^{(\gamma)}(\beta)$ for every
    $\gamma \in [0,1]$.
    Since the output itself belongs to the relaxed benchmark, its error on
    every target distribution is at most the maximum error over that
    benchmark, which proves \eqref{eq:error-goal-relaxed}.
\end{proof}

\begin{remark}
    Under majority closure, the upper bound is independent of $\gamma$. The
    dependence on $\gamma$ in Theorems~\ref{thm:lb}
    and~\ref{thm:ub} arises
    for general classes, not from the MDL argument.
\end{remark}

\end{document}